\documentclass[10pt,twocolumn,letterpaper]{article}

\usepackage[table, dvipsnames]{xcolor}
\usepackage{multirow}

\usepackage{iccv}              

\definecolor{iccvblue}{rgb}{0.21,0.49,0.74}
\usepackage[pagebackref,breaklinks,colorlinks,allcolors=iccvblue]{hyperref}

\def\paperID{3} 
\def\confName{ICCV}
\def\confYear{2025}

\title{Perceptual Classifiers: Detecting Generative Images using Perceptual Features}

\author{
    Krishna Srikar Durbha\footref{shared_first_author}\\
    The University of Texas at Austin\\
\and
    Asvin Kumar Venkataramanan\footref{shared_first_author}\\
    The University of Texas at Austin\\
\and
     Rajesh Sureddi\\
    The University of Texas at Austin\\
\and
    Alan C. Bovik\\
    University of Colorado Boulder\\
}

\begin{document}
\maketitle

\begin{abstract}
Image Quality Assessment (IQA) models are employed in many practical image and video processing pipelines to reduce storage, minimize transmission costs, and improve the Quality of Experience (QoE) of millions of viewers. These models are sensitive to a diverse range of image distortions and can accurately predict image quality as judged by human viewers. Recent advancements in generative models have resulted in a significant influx of ``GenAI'' content on the internet. Existing methods for detecting GenAI content have progressed significantly with improved generalization performance on images from unseen generative models. Here, we leverage the capabilities of existing IQA models, which effectively capture the manifold of real images within a bandpass statistical space, to distinguish between real and AI-generated images. We investigate the generalization ability of these perceptual classifiers to the task of GenAI image detection and evaluate their robustness against various image degradations. Our results show that a two-layer network trained on the feature space of IQA models demonstrates state-of-the-art performance in detecting fake images across generative models, while maintaining significant robustness against image degradations.
\end{abstract}
\footnotetext[1]{
\label{shared_first_author}
The authors contributed equally. This research was sponsored by grant number 2019844 from the National Science Foundation AI Institute for Foundations of Machine Learning (IFML). Correspondence to Krishna S. Durbha (krishna.durbha@utexas.edu), Asvin K. Venkataramanan (asvin@utexas.edu) \& Rajesh Sureddi (rajesh.sureddi@utexas.edu).
}

\section{Introduction}
\label{sec:introduction}
\begin{figure}[!ht]
    \centering
    \includegraphics[width=0.8\columnwidth]{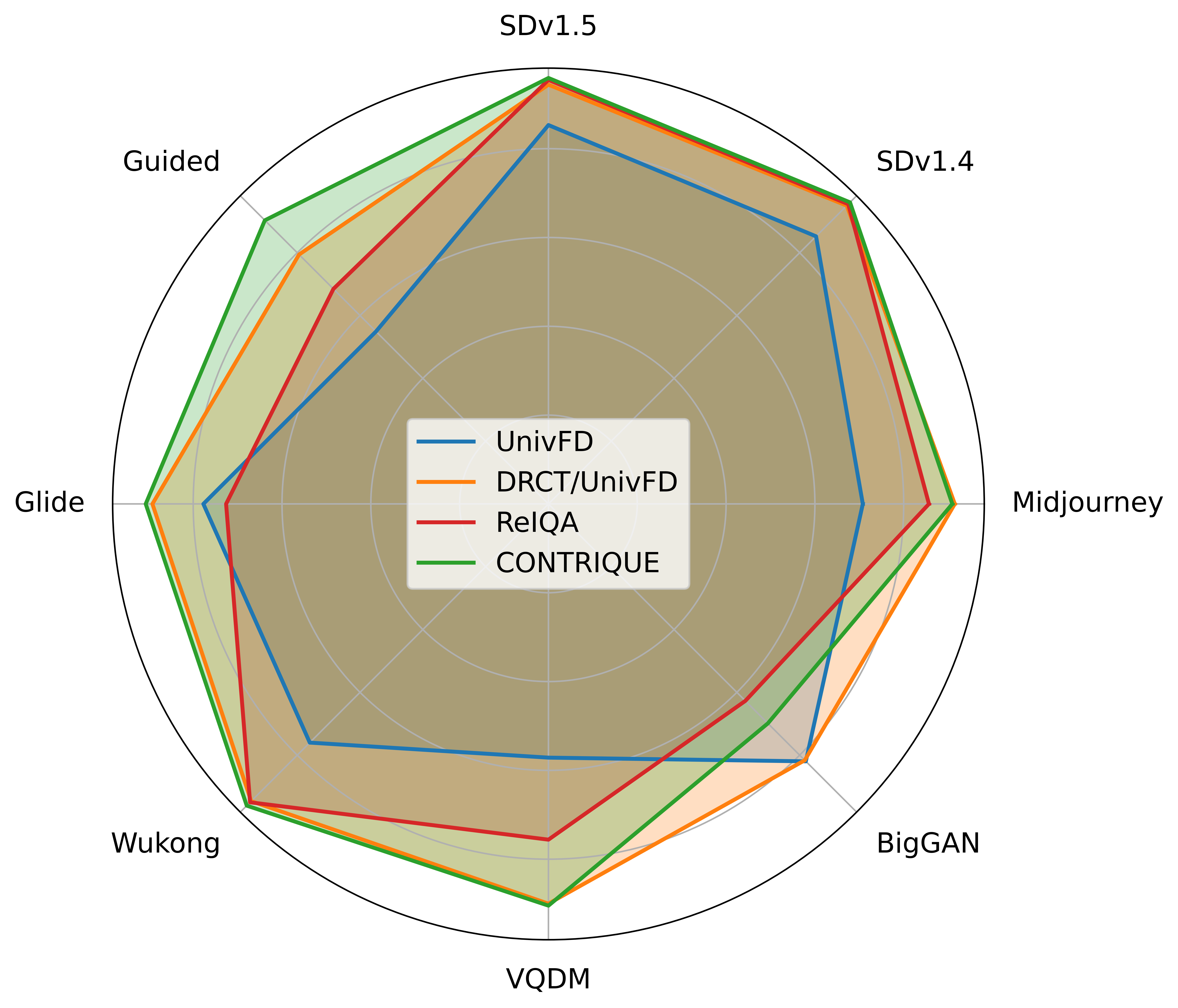}
    \caption{The generalization performance of proposed perceptual classifiers, UnivFD \cite{UnivFD} and DRCT/UnivFD \cite{DRCT}. All methods were trained and tested on the GenImage dataset \cite{GenImage}, with higher performing models appearing closer to the outer circle, indicating better performance.}
    \label{fig:Polar-Plot}
\end{figure}

Rapid advancements in generative models have revolutionized the creation and editing of images and videos. Generative models can now produce highly realistic visuals and edit images based on simple textual descriptions. As a result, the internet is inundated with AI-generated content, including images with altered faces and backgrounds, as well as malicious content that is realistic enough to challenge existing detection algorithms. Some estimates suggest that over 15 billion AI-generated images have been uploaded to the internet over the past two years \cite{everypixel}. In recent years, advancements in generative models have progressed from Generative Adversarial Networks (GAN) \cite{GAN, ProGAN, BigGAN, StyleGAN, GauGAN} to Diffusion Models (DM) \cite{Stable-Diffusion, ADM, Wukong, GLIDE}, Auto-Regressive models \cite{DALL-E}, and many others. While these advancements have deepened our understanding of image generation and enabled new creative possibilities, they also pose significant challenges to detecting and addressing visual misinformation.

Early studies on the detection of fake images (AI-generated) relied on texture patterns \cite{Detecting-fake-images-by-identifying-potential-texture-difference}, frequency analysis \cite{Freq-Spec}, co-occurrence matrices \cite{Co-Occurance}, physical scene constraints \cite{inconsistent-reflections}, compression artifacts \cite{jpeg-dimples}, and camera characteristics such as photo-response non-uniformity (PRNU) patterns \cite{prnu-detection}, among others. Although these methods were effective in detecting manipulated images, GAN-generated images, etc. their capabilities did not effectively extend to detecting images from more complex generative models. \citet{CNN-Spot} proposed a CNN-based detector trained on images generated by ProGAN \cite{ProGAN}, and showed its effectiveness at detecting images from other models in the same generative family. Despite its success, their work struggled at detecting DM-generated images \cite{On-the-detection-of-synthetic-images-generated-by-diffusion-models}. \citet{UnivFD} proposed a new fake image detection dataset built upon the ForenSynths dataset \cite{CNN-Spot}, with images added from the latest state-of-the-art generative models. To improve generalization, they proposed leveraging the visual encoder of CLIP \cite{CLIP} to extract features for fake image detection, employing basic techniques such as K-Nearest Neighbors and linear probing for classification. They demonstrated that the classification process occurs best in a feature space that has not been specifically learned to separate real and fake images. This is because a trained feature extractor easily learns patterns from generative models and treats the `real' class as a sink class, thereby reducing generalization performance. They also demonstrated that the performance of CLIP's visual encoder \cite{CLIP} surpasses that of models trained on ImageNet \cite{ImageNet} for fake image detection. The authors attribute this superior performance to CLIP's exposure to a much larger distribution of real images than models trained on ImageNet. However, a significant drawback of these classifiers is the computationally intensive nature of the CLIP visual encoder.

Concurrent with the rapid growth of visual content on the internet, image quality assessment (IQA) algorithms such as SSIM \cite{SSIM} and VMAF \cite{netflix2016_VMAF} have become integral parts of the workflows of many social media platforms and streaming services. These algorithms help reduce storage and transmission bandwidths while improving the Quality of Experience (QoE) of streamed images. These models are trained and benchmarked on datasets obtained from human studies and aim to accurately predict the quality of images containing natural and synthetic distortions. These algorithms operate by measuring deviation from naturalness caused by image degradations to predict image quality.

In this paper, we present perceptual classifiers that utilize IQA models to effectively distinguish between real and AI-generated images. Given their ubiquity in image delivery workflows, we leverage the feature space of recent SoTA no-reference IQA models (NR-IQA) for detecting fake images. We hypothesize that IQA models trained on large datasets of real images with natural and synthetic distortions implicitly model the distribution of real images in terms of image quality and distortions. Since IQA models are effective at capturing visual perturbations on real images, we leverage their feature space to train a classifier for fake image detection. Fig. \ref{fig:Polar-Plot} demonstrates the generalization performance of perceptual classifiers trained on CONTRIQUE \cite{Contrique} and ReIQA \cite{ReIQA} features against UnivFD \cite{UnivFD} and DRCT/UnivFD \cite{DRCT}. All classifiers were trained on images from the Stable Diffusion (SD) v1.4 subset of the GenImage \cite{GenImage} dataset and are evaluated on images from all generative models in the GenImage dataset. Despite not being trained on an internet-scale dataset like CLIP, perceptual classifiers achieve state-of-the-art performance on the GenImage dataset \cite{GenImage} with great generalization capabilities across unseen generative models. Additionally, perceptual classifiers employed in our experiments use CNN-based IQA backbones, which are computationally more efficient when compared to transformers and provide faster inference at scale. These approaches have the benefit of being multi-task, since the extracted features are simultaneously being used to predict image quality and to detect fake images. The following are our main contributions:
\begin{itemize}
    \setlength{\itemsep}{0.1pt}
    \item We experimented with the feature space of recent state-of-the-art no-reference image quality assessment models to train a classifier to distinguish between real and AI-generated images.
    \item We achieved state-of-the-art (SOTA) performance on multiple datasets and studied generalization capabilities on images created by unseen generative models.
    \item We evaluated the robustness of our proposed classifiers against different kinds of image degradations and validated them against recent SOTA fake image detection methods.
\end{itemize}

\nopagebreak
\section{Related Work}
\label{sec:related-works}
\begin{figure*}
    \centering
    \includegraphics[width=\linewidth]{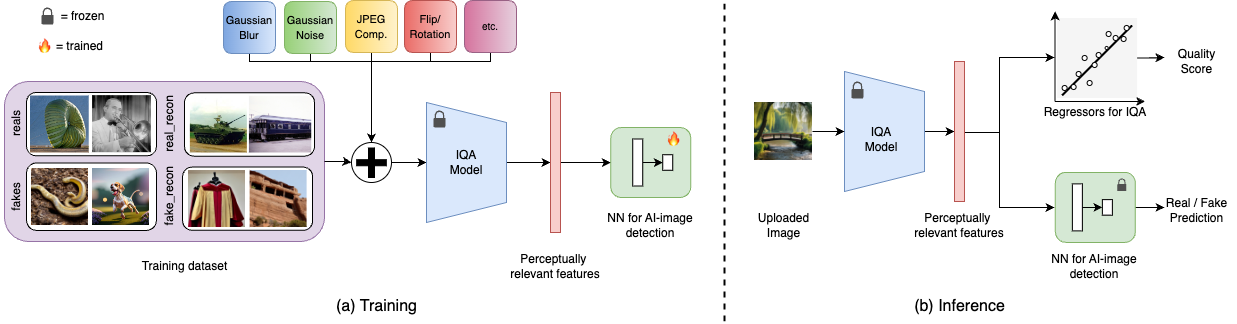}
    \caption{Overview of the training and testing procedure for IQA-based classifiers: (a) A two-layer neural network trained on perceptual features extracted by IQA models from real, fake, real reconstructed, and fake reconstructed images, along with data augmentation. (b) During inference, perceptual features extracted from uploaded images can be used to predict image quality and distinguish between real images and fakes.}
    \label{fig:training-testing}
\end{figure*}

\subsection{Fake Image Detection Methods}
\textbf{Early Approaches:} Some initial approaches to detect fake images involved matching texture patterns \cite{Detecting-fake-images-by-identifying-potential-texture-difference}, leveraging co-occurrence matrices \cite{roydetecting}, and using gram matrices \cite{liu2020global}. Multiple patch-based approaches have also been proposed that avoid processing entire images \cite{zhong2024patchcraft, chen2024single, Patch-Classifier}. Zhang \etal \cite{Detecting-and-Simulating-Artifacts-in-GAN-Fake-Images} and Frank \etal \cite{frank2020leveraging} were some of the first to discover fingerprints of GAN images in the frequency domain, which they attributed to the upsampling component in GAN pipelines. Similar grid-like Fourier fingerprints were reproduced by \cite{CNN-Spot} in their widely-used ForenSynths dataset. Multiple approaches have since been proposed to leverage these spectral artifacts to detect fake images \cite{marra2019gans,yu2019attributing, Attributing-and-Detecting-Fake-Images-Generated-by-Known-GANs, Synthbuster}. Later works \cite{UnivFD, On-the-detection-of-synthetic-images-generated-by-diffusion-models, DEFAKE,corvi2023intriguing} also investigated and confirmed that DM-generated images may contain classifiable fingerprints in their frequency spectra. However, as shown in \cite{On-the-detection-of-synthetic-images-generated-by-diffusion-models}, not all models cause grid-like Fourier patterns, suggesting poor generalization of frequency-based approaches as GenAI models advance.

\textbf{Generalization:} The generalizability of fake image detectors across datasets, as well as to unseen generative models, has been an important area of study. Some early classifiers were shown to yield poor performance on detecting images produced by different GenAI models belonging to the same family \cite{Detecting-and-Simulating-Artifacts-in-GAN-Fake-Images, cozzolino2018forensictransfer}. Wang \etal \cite{CNN-Spot} proposed a simpler approach that delivered impressive generalization performance, by fine-tuning a ResNet-50 model on real images from LSUN \cite{LSUN} and on fake images generated by 20 variants of ProGAN \cite{ProGAN}, each trained for a different LSUN category. Their work showed that detectors trained on images produced by a single GAN model can generalize to other models from the same family. However, as \cite{On-the-detection-of-synthetic-images-generated-by-diffusion-models} noted, detectors trained on GAN images do not necessarily generalize well to DM images. Ojha \etal propose an approach to tackle this issue by using an unbiased feature space. Another recent approach of note is DIRE \cite{DIRE}. The authors observed that images generated by DMs can be better reconstructed by DMs than real images. They used this observation to train a ResNet-50-based detector for distinguishing between real images and fakes. Their work leverages components key to the generation of fake images as a way of detecting them. Meanwhile, using a semantically paired dataset, \cite{DEFAKE} proposed feeding both the images and text embeddings from CLIP to a two-layer perceptron. This was motivated by their observation that images generated using text-to-image generators often fail to add background detail present in real images.

\textbf{State-of-the-Art:}
DRCT by \citet{DRCT} is the current SOTA approach and builds on top of findings of DIRE \cite{DIRE}. Their framework involves creating hard samples by reconstructing real and fake images using generative models. These reconstructed hard samples are then employed to train a classifier, using contrastive learning and cross-entropy loss functions. This procedure improves generalization performance by empowering classifiers to detect subtle traces left behind by generative models.

\subsection{Image Quality Assessment}
No reference image quality assessment (NR-IQA) focuses on predicting the mean opinion score of distorted images with no information about any pristine images. Over the years, a variety of NR-IQA models have been proposed, including BRISQUE \cite{Brisque}, NIQE \cite{Niqe}, DIIVINE \cite{Diivine}, and BLIINDS \cite{Bliinds}, which measure deviations from well-accepted models of the bandpass statistics of natural images to predict image quality. Given the rise in popularity of data-driven deep learning approaches, a variety of CNN-based methods like RAPIQUE \cite{RAPIQUE}, DB-CNN \cite{DB-CNN}, PQR \cite{PQR}, BIECON \cite{BIECON}, and PaQ-2-PiQ \cite{PaQ2PiQ} have been proposed. Vision-based transformers were also leveraged for image quality assessment in models like MUSIQ \cite{MUSIQ}, TReS \cite{TReS}, and Max-ViT \cite{maxvit}. Self-supervised contrastive learning approaches like CONTRIQUE \cite{Contrique}, ReIQA \cite{ReIQA}, and ARNIQA \cite{ARNIQA}, which emerged in response to data constraints, have proven to be powerful SOTA models with excellent generalization capabilities.

\nopagebreak
\section{Methodology}
\label{sec:methodology}
In this section, we begin by describing the framework behind perceptual classifiers, followed by a brief overview of the different IQA models we consider. Finally, we provide details on the training settings used to learn perceptual classifiers.

\subsection{Perceptual Classifier Framework}
\begin{figure}[!ht]
\centering
\begin{subfigure}[b]{0.48\columnwidth}
    \centering
    \includegraphics[width=\textwidth]{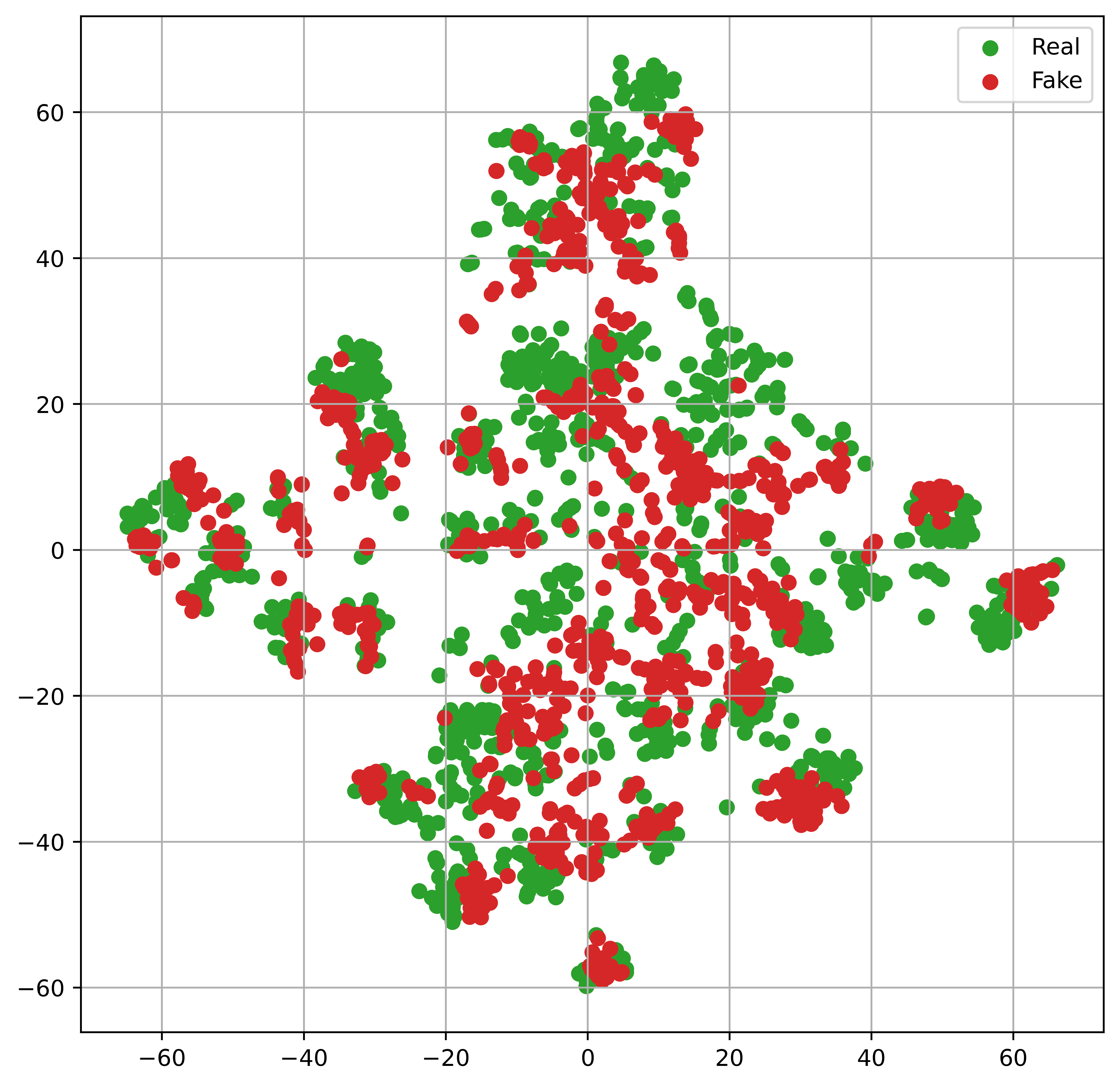}
    \caption{CLIP ViT-L/14}
    \label{M-CLIP}
\end{subfigure}
\hfill
\begin{subfigure}[b]{0.48\columnwidth}
    \centering
    \includegraphics[width=\textwidth]{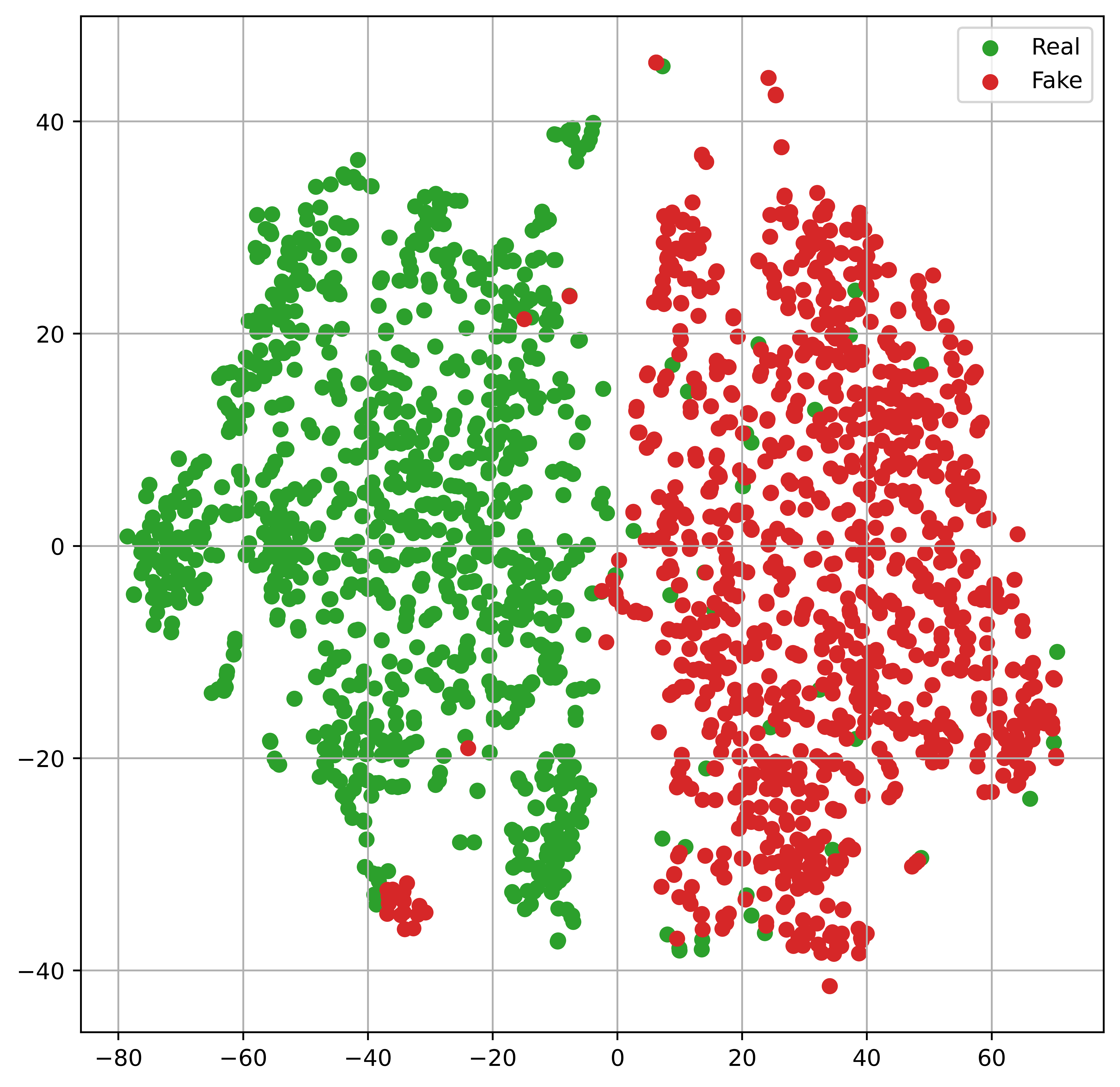}
    \caption{CONTRIQUE}
    \label{M-CONTRIQUE}
\end{subfigure}
\begin{subfigure}[b]{0.48\columnwidth}
    \centering
    \includegraphics[width=\textwidth]{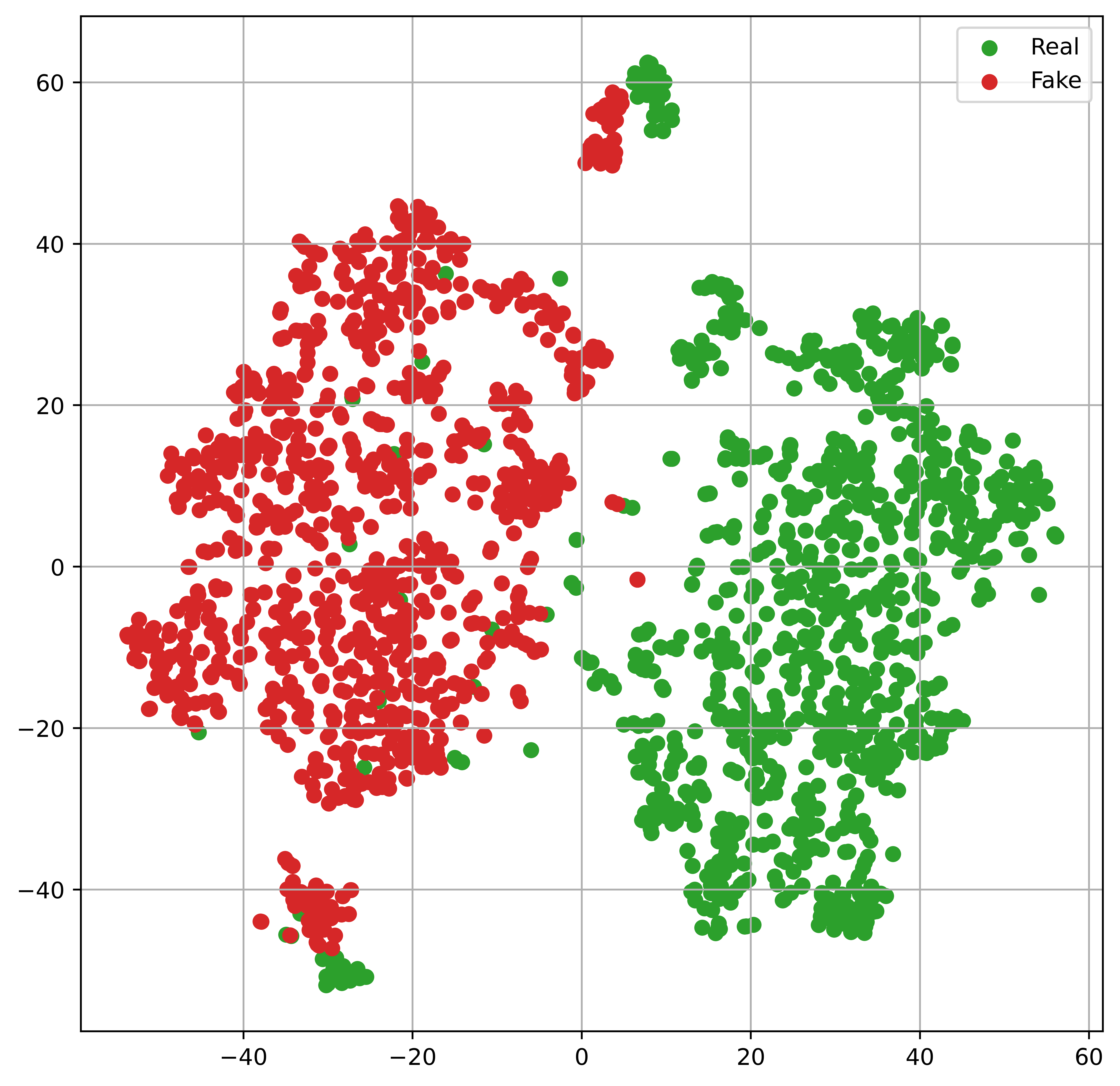}
    \caption{ReIQA}
    \label{M-ReIQA}
\end{subfigure}
\hfill
\begin{subfigure}[b]{0.48\columnwidth}
    \centering
    \includegraphics[width=\textwidth]{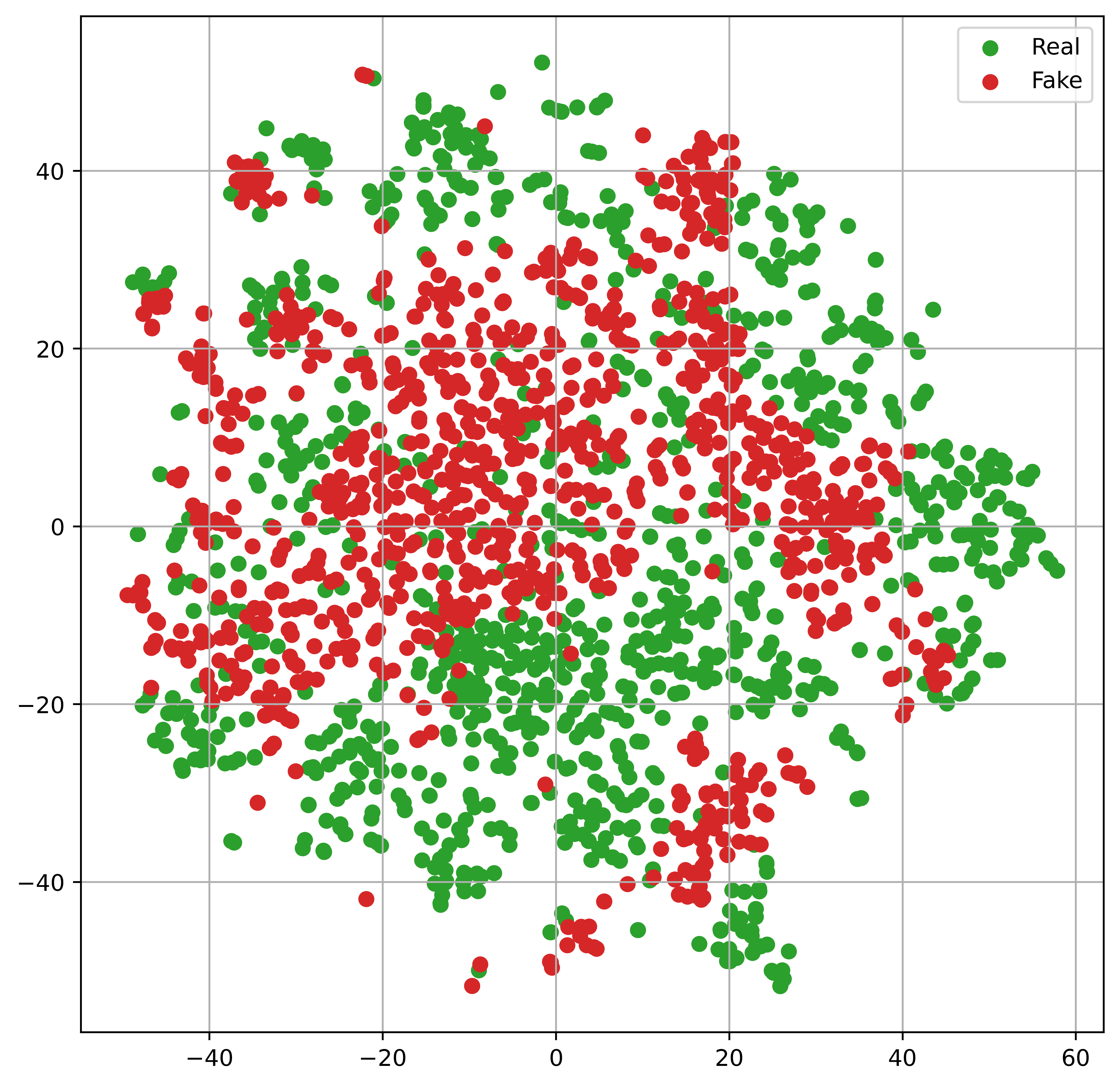}
    \caption{HyperIQA}
    \label{M-HyperIQA}
\end{subfigure}
\caption{t-SNE visualization of feature spaces of visual encoders. The feature representations are associated with real and fake images from the Stable-Diffusionv1.4 (SDv1.4) subset of the DRCT-2M dataset.}
\label{fig:feature-representations}
\end{figure}

Traditional IQA models rely on bandpass transformations such as wavelet decompositions to model the responses of visual neurons in the primary visual cortex. Recent studies, including \cite{HyperIQA, TReS, Contrique, ReIQA, ARNIQA}, have shown that features extracted from the backbones of convolutional neural networks (CNNs) possess a remarkable capacity to capture diverse perceptual artifacts. In this work, we utilize the representational strength of these perceptual features to train a classifier for detecting AI-generated images. We hypothesize that IQA models trained on large datasets of real images with various natural and synthetic distortions implicitly model the distribution of real images from a distortion perspective, thereby demonstrating an ability to separate real and fake images.

We validate our hypothesis by visualizing the feature space of various IQA models and CLIP \cite{CLIP}. We consider real and fake images from Stable Diffusion v1.4 subset in the DRCT-2M dataset. Figure \ref{fig:feature-representations} presents the t-SNE plots of feature representations extracted using the CLIP visual encoder and the backbone networks of various IQA models, including CONTRIQUE \cite{Contrique}, ReIQA \cite{ReIQA}, and HyperIQA \cite{HyperIQA}. It may be observed that the backbones of CONTRIQUE \cite{Contrique} and ReIQA \cite{ReIQA} show good separation between real images and fake images, while the representations of each class are closely grouped for CLIP \cite{CLIP} and HyperIQA \cite{HyperIQA}. These plots provide preliminary evidence of our hypothesis explaining the ability of IQA models to distinguish real and fake images.

Figure \ref{fig:training-testing} provides an overview of the training and inference frameworks. We train our perceptual classifier i.e. classifier trained on IQA features, by combining IQA features with a two-layer neural network classifier to detect real and fake images. Based on previous works \cite{DIRE, UnivFD, DRCT}, we train our classifiers on four types of image samples - real, fake, real-reconstructed, and fake-reconstructed. Similar to DRCT \cite{DRCT}, the reconstructed images are generated using the Stable Diffusion inpainting pipeline, capturing fingerprints relevant to diffusion models. The reconstructed images were generated with an empty prompt for 50 inference steps using a guidance scale of 7.5. Throughout the training process, we freeze the IQA backbone, preserving its ability to extract perceptual features. Hence, during inference, the IQA features can be directly utilized for image quality prediction using linear regressors or for real/fake image detection using a two-layer classifier. All codes associated with this work will be open-sourced.

\subsection{Image Quality Assessment Models}
In our experiments, we consider various IQA backbones, each trained using different training strategies and distortion banks. We employ state-of-the-art IQA models including HyperIQA \cite{HyperIQA}, TReS \cite{TReS}, CONTRIQUE \cite{Contrique}, ReIQA \cite{ReIQA}, and ARNIQA \cite{ARNIQA}. Among these models, HyperIQA and TReS follow a supervised training approach, i.e, they employ quality prediction scores to train their feature extractors, whereas CONTRIQUE, ReIQA, and ARNIQA employ a self-supervised training method by training their feature extractors independently of their quality prediction regressors.

HyperIQA \cite{HyperIQA} employs a hypernetwork to predict the weights of the quality prediction model using semantic features while employing global features and local distortion features to predict quality. TReS \cite{TReS} leverages the self-attention mechanism to learn a non-local image representation from multi-scale features extracted from a ResNet-50 backbone using L2 loss and relative ranking loss. Authors of CONTRIQUE \cite{Contrique} train a ResNet-50 backbone to classify different types of image distortions, ensuring the model's feature space is made sensitive to distortions without any subjective quality data. Saha et al. \cite{ReIQA} build on \cite{Contrique} by incorporating the influence of content into quality assessment, thereby capturing high-level semantic information and low-level distortion features. The authors of ARNIQA \cite{ARNIQA} train a ResNet-50 backbone on pristine images distorted exclusively using compositions of synthetic distortions. These distortion compositions are applied to different images during training to avoid any content dependency.

\begin{figure*}[!ht]
    \centering
    \includegraphics[width=0.8\linewidth]{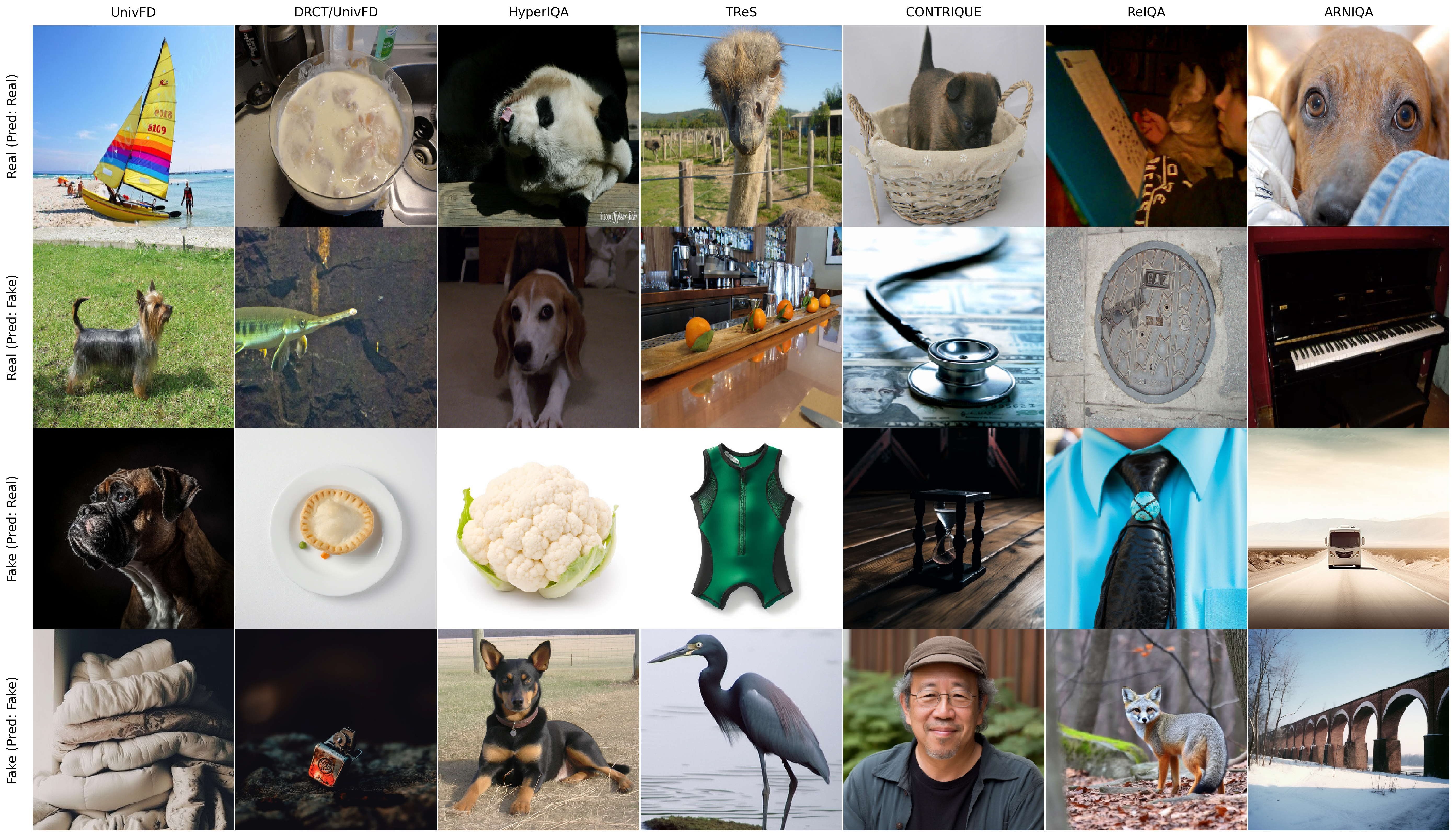}
    \caption{A visualization of predictions by multiple fake image detection models on real images (ImageNet) and fake images (Midjourney) from the GenImage dataset.}
    \label{fig:predictions}
\end{figure*}

\subsection{Training Settings}
\label{ssec:training}
We trained a two-layer neural network on perceptual features extracted from pre-trained IQA models. We employ a wide range of data augmentation techniques including Gaussian Blur with random standard deviation, JPEG Compression with a random quality factor, horizontal flip, Gaussian noise, random rotation, brightness and contrast adjustments, and grid dropout to improve robustness towards various post-processing methods. During inference, we used only data augmentation when studying specific degradation. We evaluated images at their native resolution unless the feature extractor required fixed dimensions to avoid zeroed weights from padded zeros. During training and inference, feature extractor weights were frozen for IQA models. We trained our classifiers on all images from training datasets, along with the reconstructed images generated using the SDv1 model. We trained our classifiers using a combination of margin-based contrastive loss and cross-entropy loss, as demonstrated in \cite{DRCT}. The contrastive loss minimizes the Euclidean distances between 1024-dimensional features extracted from the hidden layer of the classifier for positive pairs (similar labels) while maximizing the distances between negative pairs (different labels). The losses are calculated as follows:
\begin{align}
    \mathcal{L_{\text{CL}}} &= \sum_{i=1}^{N} \sum_{j=1}^{N} \scalebox{1.25}{$\frac{\left[y_{ij}D_{ij}^2 + (1 - y_{ij}) \max(0, m - D_{ij})^2 \right]}{N^2}$}, \\
    \mathcal{L_{\text{CE}}} &= -\frac{1}{N} \sum_{i=1}^{N} \left[ y_i \log(\hat{y}_i) + (1 - y_i) \log(1 - \hat{y}_i) \right], \\
    \mathcal{L_{\text{total}}} &= \lambda\mathcal{L_{\text{CL}}} + (1 - \lambda)\mathcal{L_{\text{CE}}}.
\end{align}
where N is the batch-size, $D_{ij}$ is the Euclidean distance between the $i^{\text{th}}$ and $j^{\text{th}}$ feature vectors, $\hat{y}_{i}$ is the predicted probability, $y_{i}$ is the target label (real or fake), and $y_{ij}$ is 1 if $y_{i} = y_{j}$, and 0 otherwise. During training, we fixed $m = 1$ and $\lambda = 0.3$ based on the analysis from \cite{DRCT}. We trained the classifiers with the AdamW optimizer with a learning-rate of $10^{-4}$ and weight-decay of $4 \times 10^{-5}$ over 20 epochs.

\nopagebreak
\section{Experiments}
\label{sec:experiments}
\subsection{Datasets}
We benchmarked the performance of the perceptual classifiers on the following datasets:
(i) \textbf{DRCT-2M}: The DRCT-2M dataset, proposed by \cite{DRCT}, contains text-to-image (T2I) and image-to-image (I2I) DM-generated images. The T2I samples were generated using prompts from the MSCOCO dataset \cite{MSCOCO} along with 10 variants of SD \cite{Stable-Diffusion}. The I2I samples were generated using 3 variants of ControlNet and 3 variants of SD-based inpainting models.
(ii) \textbf{GenImage}: The GenImage \cite{GenImage} dataset contains images generated by 7 SoTA diffusion models and one GAN model. The real images in the dataset were taken from ImageNet \cite{ImageNet}.
(iii) \textbf{UniversalFakeDetection}: The UniversalFakeDetection dataset \cite{UnivFD} contains images from various generative models. The dataset includes fake images from models  11 CNN-based generative models, 7 diffusion models, and one auto-regressive model. The real images in the UniversalFakeDetection dataset were taken from the LSUN \cite{LSUN} and LAION \cite{LAION} datasets.

\begin{table*}[!ht]
	\centering
	\tabcolsep=0.275cm
	\renewcommand{\arraystretch}{1.15}
	\resizebox{\linewidth}{!}{
	\begin{tabular}{cc cccccccc c}
		\toprule
		
		\shortstack[c]{Detection\\Method\\{}\\{}} & \shortstack[c]{Midjourney\\{}\\{}} & \shortstack[c]{Stable\\Diffusion\\v1.4} & \shortstack[c]{Stable\\Diffusion\\v1.5} & \shortstack[c]{Guided\\Diffusion\\(ADM)} & \shortstack[c]{GLIDE\\{}\\{}} & \shortstack[c]{Wukong\\{}\\{}} & \shortstack[c]{VQDM\\{}\\{}} & \shortstack[c]{BigGAN\\{}\\{}} & \shortstack[c]{mAcc(\%)\\{}\\{}}\\

		\midrule
		
		{\shortstack[c]{F3-Net}} \cite{F3-Net} &
		77.85 & 98.99 & 99.08 & 51.20 & 54.87 & 97.92 & 58.99 & 49.21 & 73.51\\

		{\shortstack[c]{GramNet}} \cite{Gram-Net} &
		73.68 & 98.85 & 98.79 & 51.52 & 55.38 & 95.38 & 55.15 & 49.41 & 72.27\\
		
		\midrule

		{\shortstack[c]{CNN-Spot}} \cite{CNN-Spot} &
		84.92 & 99.88 & 99.76 & 53.48 & 53.80 & 99.68 & 55.50 & 49.93 & 74.62 \\
		
		{\shortstack[c]{DIRE}} \cite{DIRE} &
		50.40 & \textbf{99.99} & \textbf{99.92} & 52.32 & 67.23 & \textbf{99.98} & 50.10 & 49.99 & 71.24 \\

		{\shortstack[c]{De-Fake}} \cite{DEFAKE} &
		79.88 & 98.65 & 98.62 & 71.57 & 78.05 & 98.42 & 78.31 & 74.37 & 84.73 \\

		{\shortstack[c]{CLIP/RN50}} \cite{UnivFD} &
		83.30 & \underline{99.97} & \underline{99.89} & 54.55 & 57.37 & 99.52 & 57.90 & 50.00 & 75.31 \\

		{\shortstack[c]{UnivFD}} \cite{UnivFD} &
		91.46 & 96.41 & 96.14 & 58.07 & 73.40 & 94.53 & 67.83 & 57.72 & 79.45 \\
				
		{\shortstack[c]{DRCT/ConvNext-B}} \cite{DRCT} &
		\textbf{94.63} & 99.88 & 99.82 & 61.78 & 65.92 & \underline{99.91} & 74.88 & 58.81 & 82.08 \\
		
		{\shortstack[c]{DRCT/UnivFD}} \cite{DRCT} &
		\underline{91.50} & 95.00 & 94.41 & \underline{79.42} & \underline{89.18} & 94.66 & \underline{90.02} & \textbf{81.63} & \underline{89.48} \\
		
		\midrule
    
    	\rowcolor{lightgray!30}
		{\shortstack[c]{HyperIQA}}
    	& 68.42 & 74.09 & 73.61 & 56.92 & 67.02 & 72.03 & 61.86 & 50.79 & 65.59\\

		\rowcolor{lightgray!30}
		{\shortstack[c]{TReS}}
    	& 65.4 & 70.12 & 69.82 & 52.22 & 60.4 & 69.95 & 58.25 & 33.71 & 59.98\\

		\rowcolor{lightgray!30}
		{\shortstack[c]{ARNIQA}}
    	& 80.12 & 85.64 & 85.41 & 72.39 & 69.57 & 84.7 & 71.07 & 60.2 & 76.14\\
		
		\rowcolor{lightgray!30}
		{\shortstack[c]{ReIQA}}
		& 85.68 & 95.47 & 95.49 & 68.48 & 72.59 & 95.02 & 75.6 & 62.71 & 81.38\\
		
		\rowcolor{lightgray!30}
		{\shortstack[c]{CONTRIQUE}}
		& 90.94 & 96.04 & 95.91 & \textbf{90.32} & \textbf{90.68} & 96.08 & \textbf{90.45} & \underline{69.91} & \textbf{90.04}\\    
		
		\bottomrule
	\end{tabular}}
	\caption{Performance of various real/fake image detection methods evaluated across various generative models on the GenImage test dataset using classification accuracy as the evaluation metric.}
	\label{table:default_GenImage_mAcc}
\end{table*}

We trained our classifiers on reals and fakes from the Stable-Diffusion v1.4 subset (similar to \cite{DRCT} and \cite{GenImage}) along with those reconstructed using SDv1. We test the generalization performance of the trained classifier across various unseen generative models. We trained and tested our models on the DRCT-2M and GenImage datasets while employing the UniversalFakeDetection exclusively for testing due to its significance in image classification literature.

\subsection{Evaluation Metrics and Detector Baselines}
We evaluated the detectors using mean classification accuracy (mAcc) on the GenImage and DRCT datasets, setting the threshold to 0.5 as in \cite{DRCT, GenImage}. For UniversalFakeDetection, we estimated the optimal threshold using the validation set and applied it during evaluation as per work \cite{UnivFD}.

\begin{figure}
    \centering
    \includegraphics[width=\columnwidth]{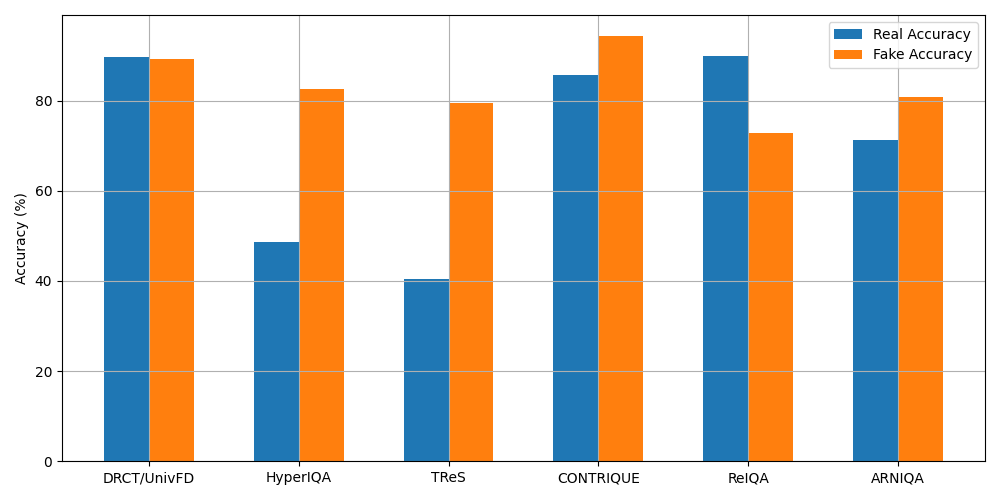}
    \caption{The mean accuracy of detecting the real and fake images on the GenImage test dataset using various fake image detection methods.}
    \label{fig:genimage_real_vs_fake_accuracy}
\end{figure}

\nopagebreak
\section{Results}
\label{sec:resuls}
Next, we discuss and compare the performance of perceptual models against SOTA models on the datasets considered. We also evaluated the robustness of our proposed methods against various image degradations.

\begin{table*}
	\centering
	\tabcolsep=0.225cm
	\renewcommand{\arraystretch}{1.15}
	\resizebox{\linewidth}{!}{
        \begin{tabular}{l cccccc cc cc ccc ccc c}
            \toprule
            
            \multirow{2}{*}{\shortstack[c]{Method}} & \multicolumn{6}{c}{\shortstack[c]{SD Variants}} & \multicolumn{2}{c}{\shortstack[c]{Turbo Variants}} & \multicolumn{2}{c}{\shortstack[c]{LCM Variants}} & \multicolumn{3}{c}{\shortstack[c]{ControlNet Variants}} & \multicolumn{3}{c}{\shortstack[c]{DR Variants}} & \multirow{2}{*}{\shortstack[c]{mAcc\\(\%)}} \\

            \cmidrule(lr){2-7} \cmidrule(lr){8-9} \cmidrule(lr){10-11} \cmidrule(lr){12-14} \cmidrule(lr){15-17}
            
            & \shortstack[c]{LDM\\{}} & \shortstack[c]{SDv1.4\\{}} & \shortstack[c]{SDv1.5\\{}} & \shortstack[c]{SDv2\\{}} & \shortstack[c]{SDXL\\{}} & \shortstack[c]{SDXL-\\Refiner} & \shortstack[c]{SD-\\Turbo} & \shortstack[c]{SDXL-\\Turbo} & \shortstack[c]{{}\\LCM-\\SDv1.5} & \shortstack[c]{LCM-\\SDXL} & \shortstack[c]{SDv1-\\Ctrl} & \shortstack[c]{SDv2-\\Ctrl} & \shortstack[c]{SDXL-\\Ctrl} & \shortstack[c]{SDv1-\\DR} & \shortstack[c]{SDv2-\\DR} & \shortstack[c]{SDXL-\\DR} & \\
            
            \midrule
            
            {\shortstack[c]{F3-Net}} \cite{F3-Net} &
            99.85 & 99.78 & 99.79 & 88.66 & 55.85 & 87.37 & 68.29 & 63.66 & 97.39 & 54.98 & 97.98 & 72.39 & 81.99 & 65.42 & 50.39 & 50.27 & 77.13 \\
            
            {\shortstack[c]{GramNet}} \cite{Gram-Net} & 
            99.40 & 99.01 & 98.84 & 95.30 & 62.63 & 80.68 & 71.19 & 69.32 & 93.05 & 57.02 & 89.97 & 75.55 & 82.68 & 51.23 & 50.01 & 50.08 & 76.62 \\
            
            {\shortstack[c]{CNNSpot}}  \cite{CNN-Spot} & 
            99.87 & 99.91 & 99.90 & \underline{97.55} & 66.25 & 86.55 &  86.15 & 72.42 & 98.26 & 61.72 & 97.96 & 85.89 & 82.84 & 60.93 & 51.41 & 50.28 & 81.12 \\

            {\shortstack[c]{DIRE}} \cite{DIRE} & 
            98.19 & \underline{99.94} & \textbf{99.96} & 68.16 & 53.84 & 71.93  & 58.87 & 54.35 & \textbf{99.78} & 59.73 & 99.65 & 64.20 & 59.13 & 51.99 & 50.04 & 49.97 & 71.23 \\

            {\shortstack[c]{De-Fake}} \citep{DEFAKE} &
            92.1 & 99.53 & 99.51 & 89.65 & 64.02 & 69.24 & 92.00 & 93.93 & 99.13 & 70.89 & 58.98 & 62.34 & 66.66 & 50.12 & 50.16 & 50.00 & 75.52 \\
            
            {\shortstack[c]{CLIP/RN50}} \cite{UnivFD} &
            \underline{99.00} & \textbf{99.99} & \textbf{99.96} & 94.61 & 62.08 & 91.43  & 83.57 & 64.40 & 98.97 & 57.43 & \underline{99.74} & 80.69 & 82.03 & 65.83 & 50.67 & 50.47 & 80.05 \\
            
            {\shortstack[c]{UnivFD}} \cite{UnivFD} &
            98.30 & 96.22 & 96.33 & 93.83 & 91.01 & \textbf{93.91} & 86.38 & 85.92 & 99.04 & \underline{88.99} & 90.41 & 81.06 & 89.06 & 51.96 & 51.03 & 50.46 & 83.46 \\
            
            {\shortstack[c]{DRCT/Conv-B}} \cite{DRCT} &
            \textbf{99.91} & 99.90 & \underline{99.90} & 96.32 & 83.87 & 85.63 & \underline{91.88} & 70.04 & \underline{99.66} & 78.76 & \textbf{99.90} & \textbf{95.01} & 81.21 & \textbf{99.90} & \textbf{95.40} & \underline{75.39} & \underline{90.79} \\
            
            {\shortstack[c]{DRCT/UnivFD}} \cite{DRCT} &
            96.74 & 96.33 & 96.33 & 94.89 & \textbf{96.24} & \underline{93.46} & \textbf{93.87} & \underline{92.94} & 91.17 & \textbf{95.01} & 93.90 & \underline{92.68} & \textbf{91.95} & 94.10 & 69.55 & 57.43 & 90.49 \\

            \midrule

            \rowcolor{lightgray!30}
            {\shortstack[c]{HyperIQA}} & 
            81.11 & 80.91 & 80.9 & 80.4 & 72.15 & 72.53 & 74.36 & 72.45 & 78.66 & 63.87 & 78.4 & 70.74 & 71.78 & 63.44 & 55.82 & 52.32 & 71.86\\
    
            \rowcolor{lightgray!30}
            {\shortstack[c]{TReS}} & 
            87.24 & 87.2 & 87.19 & 86.92 & 79.58 & 78.89 & 79.4 & 78.71 & 85.64 & 75.56 & 86.23 & 80.78 & 83.96 & 64.15 & 52.88 & 52.0 & 77.9\\
    
            \rowcolor{lightgray!30}
            {\shortstack[c]{ARNIQA}} & 
            87.21 & 87.1 & 87.03 & 86.93 & 80.62 & 82.78 & 80.73 & 81.57 & 84.12 & 72.42 & 86.3 & 79.45 & 80.67 & 80.78 & 64.33 & 51.44 & 79.59\\
    
            \rowcolor{lightgray!30}
            {\shortstack[c]{ReIQA}} & 
            96.45 & 96.23 & 96.12 & 93.8 & 77.51 & 72.43 & 80.3 & 73.88 & 91.54 & 82.47 & 94.3 & 78.91 & 80.11 & 94.02 & 67.63 & 57.26 & 83.31\\
    
            \rowcolor{lightgray!30}
            {\shortstack[c]{CONTRIQUE}} & 
            98.62 & 98.57 & 98.6 & \textbf{97.72} & \underline{93.62} & 87.06 & 90.96 & \textbf{93.51} & 96.61 & 87.7 & 92.11 & 89.04 & \underline{85.79} & \underline{97.78} & \underline{71.49} & \textbf{80.55} & \textbf{91.23}\\

            \bottomrule
        \end{tabular}
    }
    \caption{Performance of various real/fake image detection methods evaluated across various generative models on the DRCT-2M test dataset using classification accuracy as the evaluation metric.}
    \label{table:drct_results}
\end{table*}

\subsection{GenImage Dataset}
Table \ref{table:default_GenImage_mAcc} shows the performance of various prior methods \cite{CNN-Spot, DIRE, UnivFD, DRCT} and our proposed perceptual classifiers on the GenImage dataset. Most of the fake image detection methods exhibit high accuracy on the SDv1.4, SDv1.5, and Wukong subsets. However, it may be observed that most of the prior methods \cite{F3-Net, Gram-Net, CNN-Spot, DIRE} fail to deliver good performance on the ADM, GLIDE, VQDM, and BigGAN subsets. Conversely, most recent state-of-the-art methods exhibit good generalization capabilities on images across a wide range of generative models. Most detection methods deliver low performance when tested on GAN-based fake images generated by BigGAN as compared to others.

It may be observed that the perceptual classifier trained on the feature space of CONTRIQUE \cite{Contrique} achieved state-of-the-art performance, surpassing its predecessors - DRCT/UnivFD, DRCT/CovnNext-B \cite{DRCT}, and UnivFD \cite{UnivFD}. CONTRIQUE outperformed the previous state-of-the-art method, DRCT/UnivFD, on six out of eight validation subsets. The differences in their performances are most significant on the ADM and BigGAN datasets, where CONTRIQUE delivers a decline in performance on BigGAN, while outperforming on ADM. The performance of the ReIQA perceptual classifier slightly trailed that of DRCT/ConvNext-B, while surpassing other prior SOTA methods, including CNN-Spot \cite{CNN-Spot}, UnivFD \cite{UnivFD}, and DIRE \cite{DIRE}. The classifier trained on features from ARNIQA \cite{ARNIQA} delivered competitive performance against methods including CNN-Spot \cite{CNN-Spot}, and DIRE \cite{DIRE}. Whereas, the classifiers trained on HyperIQA \cite{HyperIQA} and TReS \cite{TReS} showed the least performance.

The performance disparity among IQA models can be attributed to various factors, primarily training strategies and distortion banks used for training. Figure \ref{fig:genimage_real_vs_fake_accuracy} plots bar graphs showing the mean accuracy of detecting real and fake images in the GenImage \cite{GenImage} test dataset for different detection methods. Most detectors showed similar accuracies at detecting reals and fakes. Interestingly, we observed that classifiers trained on HyperIQA and TReS exhibited low accuracy on real images and high accuracy on fake images, resulting in an overall decrease in performance. Based on these results, we infer that models employing self-supervised learning like CONTRIQUE, ReIQA, and ARNIQA perform better with higher classification accuracies, while supervised learning methods like HyperIQA and TReS struggle to show good generalization performance. ARNIQA, which is only trained on images with synthetic distortions, showed slightly lower performance when compared to CONTRIQUE and ReIQA, which have also been trained on distortions present in natural images. This behavior is likely due to ARNIQA's lack of exposure to images with natural distortions. The difference in performance between CONTRIQUE and ReIQA classifiers can be attributed to their differences in training strategies used during pretraining and network architectures.

\begin{figure*}[!ht]
    \centering
    \begin{subfigure}[b]{0.24\textwidth}
        \centering
        \includegraphics[width=\textwidth]{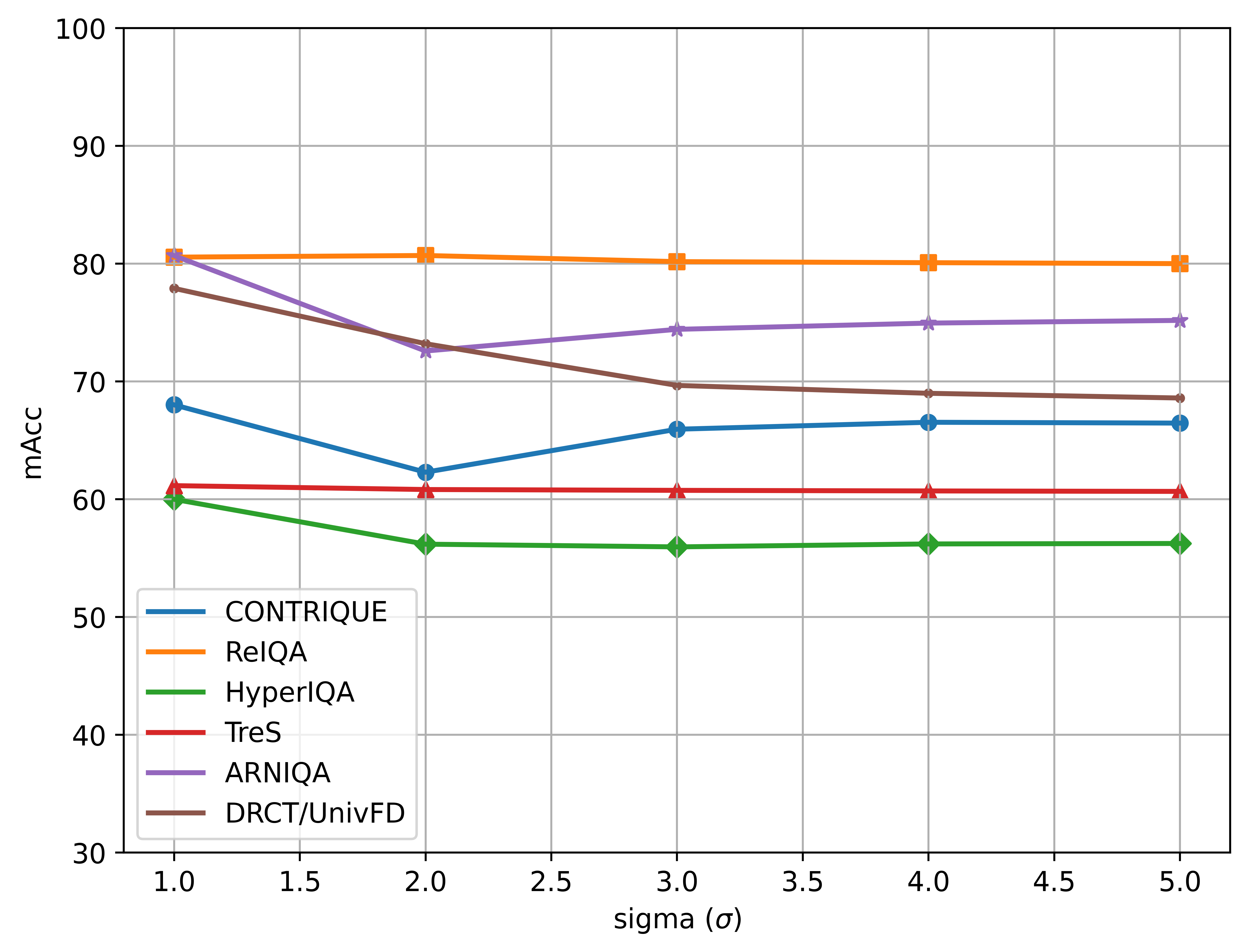}
        \caption{Gaussian Blur\\DRCT-2M}
    \end{subfigure}
    \hfill
    \begin{subfigure}[b]{0.24\textwidth}
        \centering
        \includegraphics[width=\textwidth]{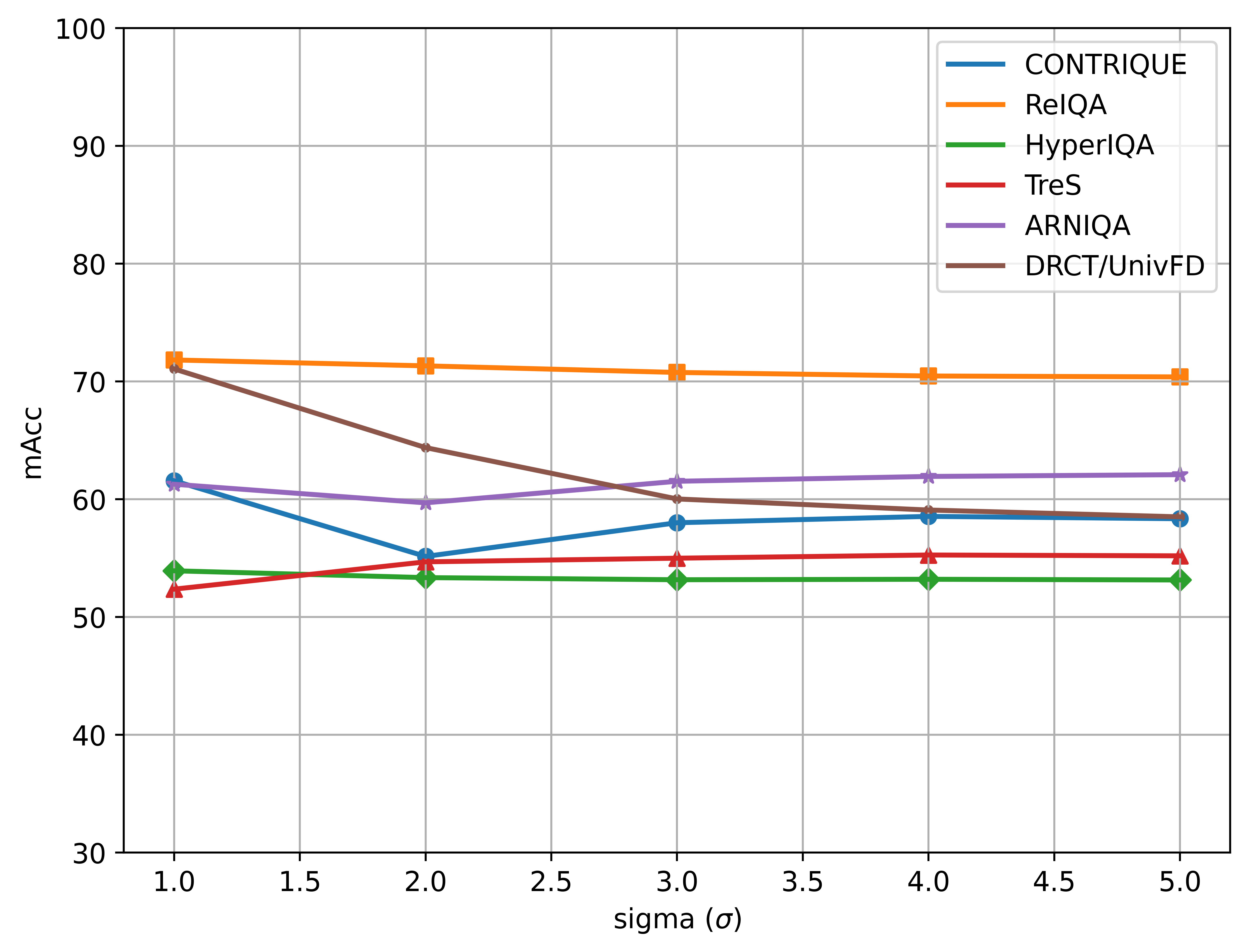}
        \caption{Gaussian Blur\\GenImage}
    \end{subfigure}
    \hfill
    \begin{subfigure}[b]{0.24\textwidth}
        \centering
        \includegraphics[width=\textwidth]{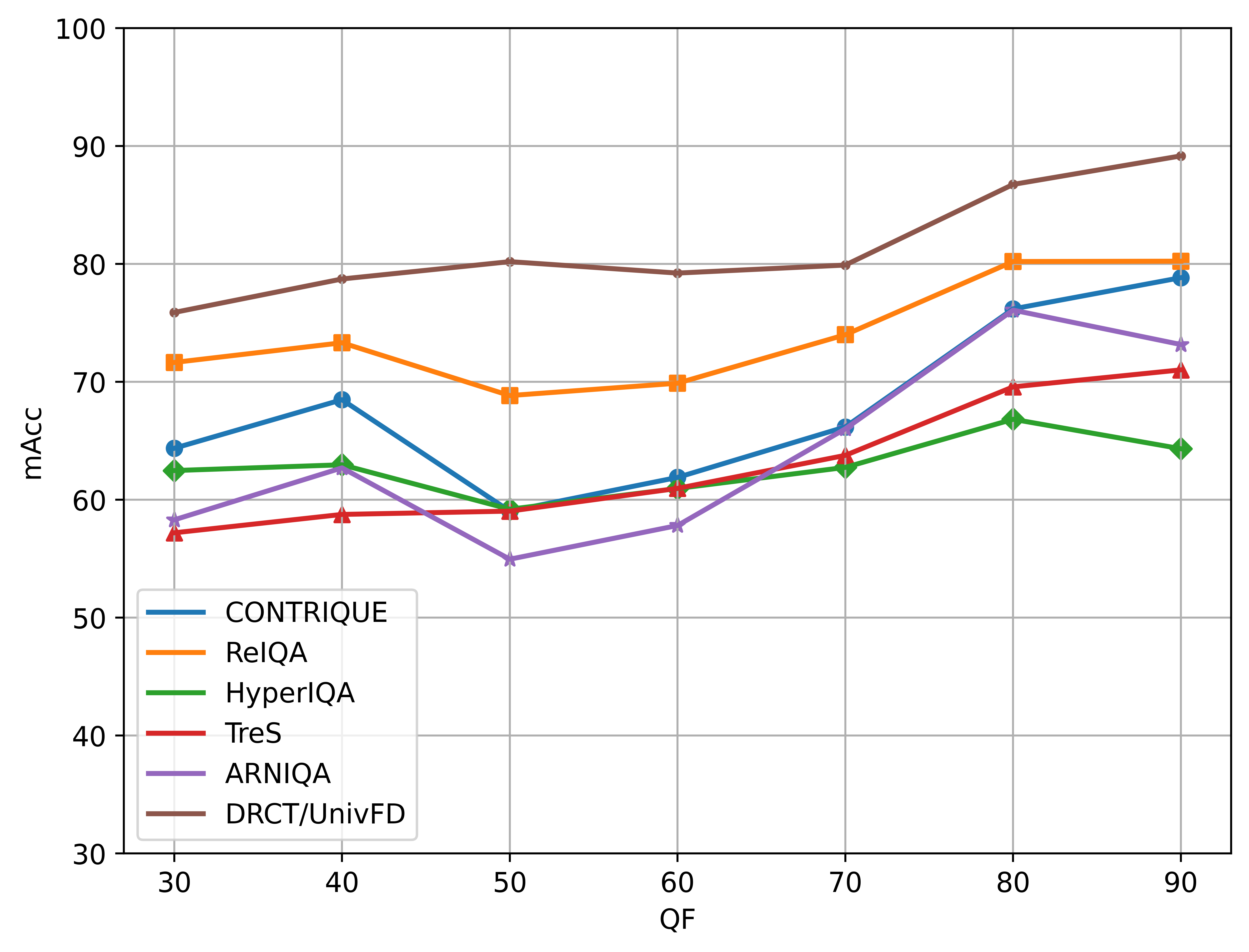}
        \caption{JPEG Compression\\DRCT-2M}
    \end{subfigure}
    \hfill
    \begin{subfigure}[b]{0.24\textwidth}
        \centering
        \includegraphics[width=\textwidth]{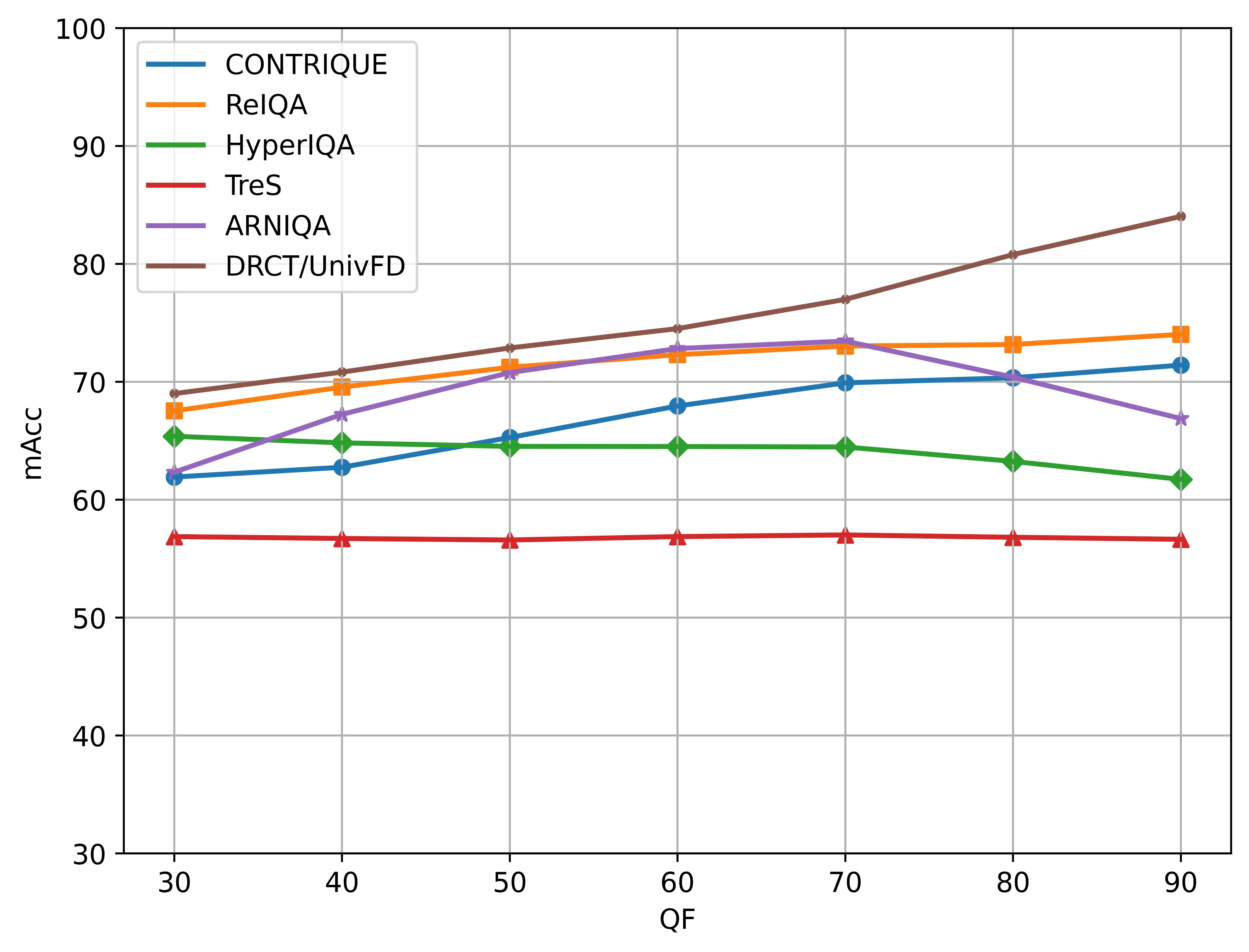}
        \caption{JPEG Compression\\GenImage}
    \end{subfigure}
    \hfill
    \caption{Mean accuracy of compared classifiers in the presence of different levels of Gaussian Blur and JPEG Compression on DRCT-2M and GenImage datasets respectively.}
    \label{fig:degradation-analysis}
\end{figure*}

\subsection{DRCT-2M Dataset}
Table \ref{table:drct_results} shows the performances of prior methods \cite{CNN-Spot, DIRE, UnivFD, DRCT} and our proposed perceptual classifiers on the DRCT-2M dataset. Similar to the performance on the GenImage dataset, it may be observed that most methods demonstrate excellent performance on subsets similar to SDv1.4, such as LDM, SDv1.5, LCM-SDv1.5, and SDv1. However, most of the prior methods \cite{F3-Net, Gram-Net, CNN-Spot, DIRE, UnivFD} suffered significant challenges in achieving high accuracy on images from unseen generative models, particularly DR variants and some SD-XL variants. The perceptual classifier trained on the CONTRIQUE \cite{Contrique} feature space achieved state-of-the-art performance once again, surpassing its predecessors - DRCT/UnivFD, DRCT/CovnNext-B \cite{DRCT}, and UnivFD \cite{UnivFD}. CONTRIQUE outperformed the previous state-of-the-art method, DRCT/UnivFD, on 9 out of 16 validation subsets. Specifically, CONTRIQUE exhibited a performance advantage over DRCT/UnivFD when evaluated on diffusion-reconstructed images (DR Variants).

It can be observed that the performance of the ReIQA \cite{ReIQA} classifier is close to the SoTA method UnivFD while trailing behind DRCT/ConvNext-B. The classifiers trained on HyperIQA \cite{HyperIQA}, TReS \cite{TReS}, and ARNIQA \cite{ARNIQA} fall behind CNN-Spot \cite{CNN-Spot} while surpassing DIRE \cite{DIRE}. Similar to the performance on the GenImage dataset, the classifiers trained on features of IQA backbones trained using self-supervised learning showed better performance than the ones trained on supervised learning. The difference in performance on the GenImage and DRCT datasets lies in their composition. The DRCT-2M dataset is entirely comprised of Stable-Diffusion variants, and the validation set contains the same set of real images from the MSCOCO dataset across all generative models. Fig. \ref{fig:predictions} shows a visualization of target label predictions by multiple fake image detectors on real images and fake images from the Midjourney subset of the GenImage dataset \cite{GenImage}.

\subsection{Cross-Dataset Performance}
We evaluated the generalization of various state-of-the-art fake image detectors trained on DRCT-2M and GenImage datasets using cross-dataset evaluation with DRCT-2M, GenImage, and UniversalFakeDetection datasets. All three datasets have different sources of real images: MSCOCO \cite{MSCOCO}, ImageNet \cite{ImageNet}, and LAION \cite{LAION}, respectively. Table \ref{table:Cross-Dataset} shows the cross-dataset performance of DRCT/ConvNext-B, DRCT/UnivFD, and our proposed perceptual classifiers based on CONTRIQUE and ReIQA. When trained on the DRCT-2M or GenImage datasets, CONTRIQUE demonstrated the best generalization performance, followed by DRCT/UnivFD, across datasets. It can be observed that all the methods suffered a reduction in accuracy when evaluated on the UniversalFakeDetection dataset, which primarily consists of fake images from multiple GANs. Among the compared models, CONTRIQUE demonstrated the best results on the UniversalFakeDetection dataset when it was trained on the GenImage dataset. These results demonstrate the competitive performance of the lighter CNN-based CONTRIQUE and ReIQA backbones as compared to the heavy CLIP:ViT-L/14's transformer-based backbone.


\begin{table}
    \centering
    \tabcolsep=0.125cm
    \renewcommand{\arraystretch}{1.05}
    \resizebox{\columnwidth}{!}{
    \begin{tabular}{l ccc ccc}
        \toprule
        
        \multirow{2}{*}{\shortstack[c]{Method}} & \multicolumn{3}{c}{\shortstack[c]{Trained on DRCT-2M}} & \multicolumn{3}{c}{\shortstack[c]{Trained on GenImage}}\\

        \cmidrule(lr){2-4} \cmidrule(lr){5-7}
        
        & \shortstack[c]{{}\\DRCT-2M\\{}} & \shortstack[c]{{}\\GenImage\\{}} & \shortstack[c]{Universal\\Fake\\Detection} & \shortstack[c]{{}\\DRCT-2M\\{}} & \shortstack[c]{{}\\GenImage\\{}} & \shortstack[c]{Universal\\Fake\\Detection} \\

        \midrule
        
        {\shortstack[c]{DRCT/Conv-B}} & 
        \underline{90.79} & 83.53 & 69.43 & 83.86 & 82.08 & 69.3 \\
        
        {\shortstack[c]{DRCT/UnivFD}} & 
        90.48 & \underline{87.67} & \textbf{75.63} & \underline{85.18} & \underline{89.49} & \underline{76.77} \\

        \midrule

        \rowcolor{lightgray!30}
        {\shortstack[c]{ReIQA}} & 
        83.31 & 79.96 & 63.79 & 81.64 & 81.38 & 69.71 \\ 

        \rowcolor{lightgray!30}
        {\shortstack[c]{CONTRIQUE}} & 
        \textbf{91.23} & \textbf{89.11} & \underline{73.52} & \textbf{85.51} & \textbf{90.04} & \textbf{79.46} \\

        \bottomrule
    \end{tabular}}
    \caption{Cross-dataset performance of various SoTA fake image detection methods and proposed classifiers trained and on different image detection datasets.}
    \label{table:Cross-Dataset}
\end{table}

\subsection{Robustness to Distortions}
Figure \ref{fig:degradation-analysis} shows mean accuracy of classifiers evaluated under varying levels of image degradations, including Gaussian blur with standard deviations 1, 2, 3, 4, and 5 (pixels); and JPEG compression with quality factors 90, 80, 70, 60, 50, 40, and 30.  Unlike the performances on datasets without distortions, the perceptual classifier trained on the CONTRIQUE features exhibited higher vulnerability to image distortions when compared to ReIQA and DRCT/UnivFD classifiers. ReIQA demonstrated superior robustness against different levels of Gaussian Blur. However, its performance on images distorted by JPEG compression was inferior compared to DRCT/UnivFD \cite{DRCT}. Among the remaining IQA classifiers, ARNIQA \cite{ARNIQA} features showed better robustness than models trained with HyperIQA \cite{HyperIQA} or TReS \cite{TReS} features, with their performance often falling between the CONTRIQUE and ReIQA classifiers. Since IQA models are trained to measure perceptual deviations from naturalness, images with common degradations pose a slight challenge to these perceptual classifiers when distinguishing between real and fake images.

\nopagebreak
\section{Conclusion}
\label{sec:conclusion}
We explored and demonstrated the effectiveness of perceptual features for detecting AI-generated images by training perceptual classifiers on feature representations from leading image quality assessment models. Our experiments showed that classifiers trained on feature spaces from IQA models generalize well to images from unseen generative models, owing to their ability to capture the distributions of real images. We achieved state-of-the-art performance on the GenImage and DRCT-2M datasets using IQA backbones trained using self-supervised learning. The cross-dataset analysis demonstrates significant improvements in generalization performance on unseen models when compared against the SOTA methods - DRCT/UnivFD and DRCT/ConvNext. Our proposed models delivered better robustness against Gaussian blur, albeit a decrease in robustness against post-processing techniques like JPEG compression compared to SOTA methods. Unlike models like CLIP, which require pre-training on internet-scale datasets, the compared IQA models are CNN-based and are trained on smaller datasets. These models are computationally efficient, since the same feature representations can be used for both quality estimation and fake image detection. Considering the ubiquity of IQA algorithms on image-hosting websites, we believe that our approach makes it possible to leverage their computation towards predicting image quality and detecting AI-generated images. In the future, we aim to expand our work to consider a bigger set of IQA backbones and images from recent high-quality image generative models.


{
    \small
    \bibliographystyle{ieeenat_fullname}
    \bibliography{main}

\begin{thebibliography}{62}
\providecommand{\natexlab}[1]{#1}
\providecommand{\url}[1]{\texttt{#1}}
\expandafter\ifx\csname urlstyle\endcsname\relax
  \providecommand{\doi}[1]{doi: #1}\else
  \providecommand{\doi}{doi: \begingroup \urlstyle{rm}\Url}\fi

\bibitem[Wuk()]{Wukong}
{Wukong}.
\newblock \url{https://xihe.mindspore.cn/modelzoo/wukong}.

\bibitem[eve(2024)]{everypixel}
{Everypixel Journal - Your Guide to the Entangled World of AI 2024}, 2024.

\bibitem[Agarwal and Farid(2017)]{jpeg-dimples}
Shruti Agarwal and Hany Farid.
\newblock {Photo forensics from JPEG dimples}.
\newblock \emph{{IEEE Workshop on Information Forensics and Security (WIFS)}}, pages 1--6, 2017.

\bibitem[Agnolucci et~al.(2024)Agnolucci, Galteri, Bertini, and Del~Bimbo]{ARNIQA}
Lorenzo Agnolucci, Leonardo Galteri, Marco Bertini, and Alberto Del~Bimbo.
\newblock {ARNIQA: Learning Distortion Manifold for Image Quality Assessment}.
\newblock \emph{{IEEE/CVF Winter Conference on Applications of Computer Vision (WACV)}}, pages {188--197}, 2024.

\bibitem[Bammey(2024)]{Synthbuster}
Quentin Bammey.
\newblock {Synthbuster: Towards Detection of Diffusion Model Generated Images}.
\newblock \emph{{IEEE Open Journal of Signal Processing}}, 5:\penalty0 1--9, 2024.

\bibitem[Brock et~al.(2018)Brock, Donahue, and Simonyan]{BigGAN}
Andrew Brock, Jeff Donahue, and Karen Simonyan.
\newblock {Large Scale GAN Training for High Fidelity Natural Image Synthesis}.
\newblock \emph{{ArXiv}}, abs/1809.11096, 2018.

\bibitem[Chai et~al.(2020)Chai, Bau, Lim, and Isola]{Patch-Classifier}
Lucy Chai, David Bau, Ser-Nam Lim, and Phillip Isola.
\newblock {What makes fake images detectable? Understanding properties that generalize}.
\newblock \emph{{European Conference on Computer Vision}}, 2020.

\bibitem[Chen et~al.(2024{\natexlab{a}})Chen, Zeng, Yang, and Yang]{DRCT}
Baoying Chen, Jishen Zeng, Jianquan Yang, and Rui Yang.
\newblock {DRCT: Diffusion Reconstruction Contrastive Training towards Universal Detection of Diffusion Generated Images}.
\newblock \emph{{International Conference on Machine Learning}}, 2024{\natexlab{a}}.

\bibitem[Chen et~al.(2024{\natexlab{b}})Chen, Yao, and Niu]{chen2024single}
Jiaxuan Chen, Jieteng Yao, and Li Niu.
\newblock {A single simple patch is all you need for AI-generated image detection}.
\newblock \emph{{arXiv preprint arXiv:2402.01123}}, 2024{\natexlab{b}}.

\bibitem[Chierchia et~al.(2014)Chierchia, Poggi, Sansone, and Verdoliva]{prnu-detection}
Giovanni Chierchia, Giovanni Poggi, Carlo Sansone, and Luisa Verdoliva.
\newblock {A Bayesian-{MRF} approach for {PRNU}-based image forgery detection}.
\newblock \emph{{IEEE Transactions on Information Forensics and Security}}, 9\penalty0 (4):\penalty0 554--567, 2014.

\bibitem[Corvi et~al.(2022)Corvi, Cozzolino, Zingarini, Poggi, Nagano, and Verdoliva]{On-the-detection-of-synthetic-images-generated-by-diffusion-models}
Riccardo Corvi, Davide Cozzolino, Giada Zingarini, Giovanni Poggi, Koki Nagano, and Luisa Verdoliva.
\newblock {On the detection of synthetic images generated by diffusion models}, 2022.

\bibitem[Corvi et~al.(2023)Corvi, Cozzolino, Poggi, Nagano, and Verdoliva]{corvi2023intriguing}
Riccardo Corvi, Davide Cozzolino, Giovanni Poggi, Koki Nagano, and Luisa Verdoliva.
\newblock { Intriguing properties of synthetic images: from generative adversarial networks to diffusion models }.
\newblock \emph{{ IEEE/CVF Conference on Computer Vision and Pattern Recognition }}, pages 973--982, 2023.

\bibitem[Cozzolino et~al.(2018)Cozzolino, Thies, R{\"o}ssler, Riess, Nie{\ss}ner, and Verdoliva]{cozzolino2018forensictransfer}
Davide Cozzolino, Justus Thies, Andreas R{\"o}ssler, Christian Riess, Matthias Nie{\ss}ner, and Luisa Verdoliva.
\newblock {Forensictransfer: Weakly-supervised domain adaptation for forgery detection}.
\newblock \emph{{arXiv preprint arXiv:1812.02510}}, 2018.

\bibitem[Deng et~al.(2009)Deng, Dong, Socher, Li, Li, and Fei-Fei]{ImageNet}
Jia Deng, Wei Dong, Richard Socher, Li-Jia Li, Kai Li, and Li Fei-Fei.
\newblock {ImageNet: A large-scale hierarchical image database}.
\newblock \emph{{2009 IEEE Conference on Computer Vision and Pattern Recognition}}, pages 248--255, 2009.

\bibitem[Dhariwal and Nichol(2021)]{ADM}
Prafulla Dhariwal and Alexander~Quinn Nichol.
\newblock {Diffusion Models Beat GANs on Image Synthesis}.
\newblock \emph{{ Advances in Neural Information Processing Systems}}, pages 8780--8794, 2021.

\bibitem[Frank et~al.(2020)Frank, Eisenhofer, Sch{\"o}nherr, Fischer, Kolossa, and Holz]{frank2020leveraging}
Joel Frank, Thorsten Eisenhofer, Lea Sch{\"o}nherr, Asja Fischer, Dorothea Kolossa, and Thorsten Holz.
\newblock {Leveraging Frequency Analysis for Deep Fake Image Recognition}.
\newblock \emph{{International Conference on Machine Learning}}, pages 3247--3258, 2020.

\bibitem[Golestaneh et~al.(2022)Golestaneh, Dadsetan, and Kitani]{TReS}
S.~Alireza Golestaneh, Saba Dadsetan, and Kris~M. Kitani.
\newblock { No-Reference Image Quality Assessment via Transformers, Relative Ranking, and Self-Consistency }.
\newblock \emph{{ {IEEE/CVF} Winter Conference on Applications of Computer Vision, {WACV} 2022, Waikoloa, HI, USA, January 3-8, 2022 }}, pages 3989--3999, 2022.

\bibitem[Goodfellow et~al.(2014)Goodfellow, Pouget-Abadie, Mirza, Xu, Warde-Farley, Ozair, Courville, and Bengio]{GAN}
Ian Goodfellow, Jean Pouget-Abadie, Mehdi Mirza, Bing Xu, David Warde-Farley, Sherjil Ozair, Aaron Courville, and Yoshua Bengio.
\newblock {Generative Adversarial Networks}.
\newblock \emph{{Advances in Neural Information Processing Systems}}, 27, 2014.

\bibitem[Joslin and Hao(2020)]{Attributing-and-Detecting-Fake-Images-Generated-by-Known-GANs}
Matthew Joslin and Shuang Hao.
\newblock {Attributing and Detecting Fake Images Generated by Known GANs}.
\newblock \emph{{ 2020 {IEEE} Security and Privacy Workshops, {SP} Workshops, San Francisco, CA, USA, May 21, 2020 }}, pages 8--14, 2020.

\bibitem[Karras et~al.(2017)Karras, Aila, Laine, and Lehtinen]{ProGAN}
Tero Karras, Timo Aila, Samuli Laine, and Jaakko Lehtinen.
\newblock {Progressive growing of gans for improved quality, stability, and variation}.
\newblock \emph{{arXiv preprint arXiv:1710.10196}}, 2017.

\bibitem[Karras et~al.(2019)Karras, Laine, and Aila]{StyleGAN}
Tero Karras, Samuli Laine, and Timo Aila.
\newblock {A Style-Based Generator Architecture for Generative Adversarial Networks}.
\newblock \emph{{ IEEE/CVF Conference on Computer Vision and Pattern Recognition }}, pages 4401--4410, 2019.

\bibitem[Ke et~al.(2021)Ke, Wang, Wang, Milanfar, and Yang]{MUSIQ}
Junjie Ke, Qifei Wang, Yilin Wang, Peyman Milanfar, and Feng Yang.
\newblock {Musiq: Multi-scale image quality transformer}.
\newblock \emph{{IEEE/CVF International Conference on Computer Vision}}, pages 5148--5157, 2021.

\bibitem[Kim and Lee(2016)]{BIECON}
Jongyoo Kim and Sanghoon Lee.
\newblock {Fully deep blind image quality predictor}.
\newblock \emph{{IEEE Journal of selected Topics in Signal Processing}}, 11\penalty0 (1):\penalty0 206--220, 2016.

\bibitem[Lin et~al.(2014)Lin, Maire, Belongie, Hays, Perona, Ramanan, Doll{\'a}r, and Zitnick]{MSCOCO}
Tsung-Yi Lin, Michael Maire, Serge~J. Belongie, James Hays, Pietro Perona, Deva Ramanan, Piotr Doll{\'a}r, and C.~Lawrence Zitnick.
\newblock Microsoft coco: Common objects in context.
\newblock 2014.

\bibitem[Liu et~al.(2020{\natexlab{a}})Liu, Qi, and Torr]{Gram-Net}
Zhengzhe Liu, Xiaojuan Qi, and Philip~HS Torr.
\newblock {Global Texture Enhancement for Fake Face Detection in the Wild}.
\newblock \emph{{ IEEE/CVF Conference on Computer Vision and Pattern Recognition }}, pages 8060--8069, 2020{\natexlab{a}}.

\bibitem[Liu et~al.(2020{\natexlab{b}})Liu, Qi, and Torr]{liu2020global}
Zhengzhe Liu, Xiaojuan Qi, and Philip~HS Torr.
\newblock {Global Texture Enhancement for Fake Face Detection in the Wild}.
\newblock \emph{{ IEEE/CVF Conference on Computer Vision and Pattern Recognition }}, pages 8060--8069, 2020{\natexlab{b}}.

\bibitem[Madhusudana et~al.(2022)Madhusudana, Birkbeck, Wang, Adsumilli, and Bovik]{Contrique}
Pavan~C Madhusudana, Neil Birkbeck, Yilin Wang, Balu Adsumilli, and Alan~C Bovik.
\newblock {Image Quality Assessment using Contrastive Learning}.
\newblock \emph{{IEEE Transactions on Image Processing}}, 31:\penalty0 4149--4161, 2022.

\bibitem[Marra et~al.(2019)Marra, Gragnaniello, Verdoliva, and Poggi]{marra2019gans}
Francesco Marra, Diego Gragnaniello, Luisa Verdoliva, and Giovanni Poggi.
\newblock {Do GANs leave artificial fingerprints?}
\newblock \emph{{ 2019 IEEE conference on multimedia information processing and retrieval (MIPR) }}, pages 506--511, 2019.

\bibitem[Mittal et~al.(2012)Mittal, Moorthy, and Bovik]{Brisque}
Anish Mittal, Anush~Krishna Moorthy, and Alan~Conrad Bovik.
\newblock {No-Reference Image Quality Assessment in the Spatial Domain}.
\newblock \emph{{IEEE Transactions on Image Processing}}, 21\penalty0 (12):\penalty0 4695--4708, 2012.

\bibitem[Mittal et~al.(2013)Mittal, Soundararajan, and Bovik]{Niqe}
Anish Mittal, Rajiv Soundararajan, and Alan~C. Bovik.
\newblock {Making a ``Completely Blind'' Image Quality Analyzer}.
\newblock \emph{{IEEE Signal Processing Letters}}, 20\penalty0 (3):\penalty0 209--212, 2013.

\bibitem[Moorthy and Bovik(2011)]{Diivine}
Anush~Krishna Moorthy and Alan~Conrad Bovik.
\newblock { Blind Image Quality Assessment: From Natural Scene Statistics to Perceptual Quality }.
\newblock \emph{{IEEE Transactions on Image Processing}}, 20\penalty0 (12):\penalty0 3350--3364, 2011.

\bibitem[Nataraj et~al.(2019{\natexlab{a}})Nataraj, Mohammed, Chandrasekaran, Flenner, Bappy, Roy-Chowdhury, and Manjunath]{Co-Occurance}
Lakshmanan Nataraj, Tajuddin~Manhar Mohammed, Shivkumar Chandrasekaran, Arjuna Flenner, Jawadul~H Bappy, Amit~K Roy-Chowdhury, and BS Manjunath.
\newblock {Detecting GAN generated fake images using co-occurrence matrices}.
\newblock \emph{{arXiv preprint arXiv:1903.06836}}, 2019{\natexlab{a}}.

\bibitem[Nataraj et~al.(2019{\natexlab{b}})Nataraj, Mohammed, Chandrasekaran, Flenner, Bappy, Roy-Chowdhury, and Manjunath]{roydetecting}
Lakshmanan Nataraj, Tajuddin~Manhar Mohammed, Shivkumar Chandrasekaran, Arjuna Flenner, Jawadul~H. Bappy, Amit~K. Roy-Chowdhury, and B.~S. Manjunath.
\newblock {Detecting GAN generated Fake Images using Co-occurrence Matrices}, 2019{\natexlab{b}}.

\bibitem[{Netflix Technology Blog}(2016)]{netflix2016_VMAF}
{Netflix Technology Blog}.
\newblock {Toward a Practical Perceptual Video Quality Metric}.
\newblock \url{https://netflixtechblog.com/toward-a-practical-perceptual-video-quality-metric-653f208b9652}, 2016.
\newblock Accessed: 2024-09-07.

\bibitem[Nichol et~al.(2022)Nichol, Dhariwal, Ramesh, Shyam, Mishkin, McGrew, Sutskever, and Chen]{GLIDE}
Alexander~Quinn Nichol, Prafulla Dhariwal, Aditya Ramesh, Pranav Shyam, Pamela Mishkin, Bob McGrew, Ilya Sutskever, and Mark Chen.
\newblock {GLIDE: Towards Photorealistic Image Generation and Editing with Text-Guided Diffusion Models}.
\newblock \emph{{ International Conference on Machine Learning, {ICML} 2022, 17-23 July 2022, Baltimore, Maryland, {USA} }}, 162:\penalty0 16784--16804, 2022.

\bibitem[O'brien and Farid(2012)]{inconsistent-reflections}
James~F O'brien and Hany Farid.
\newblock {Exposing photo manipulation with inconsistent reflections.}
\newblock \emph{{ACM Trans. Graph.}}, 31\penalty0 (1):\penalty0 4--1, 2012.

\bibitem[Ojha et~al.(2023)Ojha, Li, and Lee]{UnivFD}
Utkarsh Ojha, Yuheng Li, and Yong~Jae Lee.
\newblock { Towards Universal Fake Image Detectors that Generalize Across Generative Models }.
\newblock \emph{{ {IEEE/CVF} Conference on Computer Vision and Pattern Recognition, {CVPR} 2023, Vancouver, BC, Canada, June 17-24, 2023 }}, pages 24480--24489, 2023.

\bibitem[Park et~al.(2019)Park, Liu, Wang, and Zhu]{GauGAN}
Taesung Park, Ming-Yu Liu, Ting-Chun Wang, and Jun-Yan Zhu.
\newblock {Semantic Image Synthesis with Spatially-Adaptive Normalization}.
\newblock \emph{{ IEEE/CVF Conference on Computer Vision and Pattern Recognition }}, pages 2337--2346, 2019.

\bibitem[Qian et~al.(2020)Qian, Yin, Sheng, Chen, and Shao]{F3-Net}
Yuyang Qian, Guojun Yin, Lu Sheng, Zixuan Chen, and Jing Shao.
\newblock { Thinking in Frequency: Face Forgery Detection by Mining Frequency-aware Clues }.
\newblock \emph{{European Conference on Computer Vision}}, pages 86--103, 2020.

\bibitem[Radford et~al.(2021)Radford, Kim, Hallacy, Ramesh, Goh, Agarwal, Sastry, Askell, Mishkin, Clark, Krueger, and Sutskever]{CLIP}
Alec Radford, Jong~Wook Kim, Chris Hallacy, Aditya Ramesh, Gabriel Goh, Sandhini Agarwal, Girish Sastry, Amanda Askell, Pamela Mishkin, Jack Clark, Gretchen Krueger, and Ilya Sutskever.
\newblock {Learning Transferable Visual Models From Natural Language Supervision}.
\newblock \emph{{International Conference on Machine Learning}}, 139:\penalty0 8748--8763, 2021.

\bibitem[Ramesh et~al.(2021)Ramesh, Pavlov, Goh, Gray, Voss, Radford, Chen, and Sutskever]{DALL-E}
Aditya Ramesh, Mikhail Pavlov, Gabriel Goh, Scott Gray, Chelsea Voss, Alec Radford, Mark Chen, and Ilya Sutskever.
\newblock {Zero-Shot Text-to-Image Generation}, 2021.

\bibitem[Rombach et~al.(2021)Rombach, Blattmann, Lorenz, Esser, and Ommer]{Stable-Diffusion}
Robin Rombach, A. Blattmann, Dominik Lorenz, Patrick Esser, and Bj{\"o}rn Ommer.
\newblock {High-Resolution Image Synthesis with Latent Diffusion Models}.
\newblock \emph{{IEEE/CVF Conference on Computer Vision and Pattern Recognition (CVPR)}}, pages 10674--10685, 2021.

\bibitem[Saad and Bovik(2012)]{Bliinds}
Michele~A. Saad and Alan~C. Bovik.
\newblock { Blind quality assessment of videos using a model of natural scene statistics and motion coherency }.
\newblock \emph{{ 2012 Conference Record of the Forty Sixth Asilomar Conference on Signals, Systems and Computers (ASILOMAR) }}, pages 332--336, 2012.

\bibitem[Saha et~al.(2023)Saha, Mishra, and Bovik]{ReIQA}
Avinab Saha, Sandeep Mishra, and Alan~C Bovik.
\newblock {Re-IQA: Unsupervised Learning for Image Quality Assessment in the Wild}.
\newblock \emph{{ IEEE/CVF Conference on Computer Vision and Pattern Recognition }}, pages 5846--5855, 2023.

\bibitem[Schuhmann et~al.(2021)Schuhmann, Vencu, Beaumont, Kaczmarczyk, Mullis, Katta, Coombes, Jitsev, and Komatsuzaki]{LAION}
Christoph Schuhmann, Richard Vencu, Romain Beaumont, Robert Kaczmarczyk, Clayton Mullis, Aarush Katta, Theo Coombes, Jenia Jitsev, and Aran Komatsuzaki.
\newblock {LAION-400M: Open Dataset of CLIP-Filtered 400 Million Image-Text Pairs}.
\newblock \emph{{arXiv preprint arXiv:2111.02114}}, 2021.

\bibitem[Sha et~al.(2023)Sha, Li, Yu, and Zhang]{DEFAKE}
Zeyang Sha, Zheng Li, Ning Yu, and Yang Zhang.
\newblock { DE-FAKE: Detection and Attribution of Fake Images Generated by Text-to-Image Generation Models}.
\newblock \emph{{ ACM SIGSAC Conference on Computer and Communications Security }}, pages 3418--3432, 2023.

\bibitem[Song and Xiao(2015)]{LSUN}
Fisher Yu Yinda Zhang~Shuran Song and Ari Seff~Jianxiong Xiao.
\newblock { Lsun: Construction of a large-scale image dataset using deep learning with humans in the loop }.
\newblock \emph{{arXiv preprint arXiv:1506.03365}}, 2015.

\bibitem[Su et~al.(2020)Su, Yan, Zhu, Zhang, Ge, Sun, and Zhang]{HyperIQA}
Shaolin Su, Qingsen Yan, Yu Zhu, Cheng Zhang, Xin Ge, Jinqiu Sun, and Yanning Zhang.
\newblock { Blindly Assess Image Quality in the Wild Guided by a Self-Adaptive Hyper Network }.
\newblock \emph{{ 2020 {IEEE/CVF} Conference on Computer Vision and Pattern Recognition, {CVPR} 2020, Seattle, WA, USA, June 13-19, 2020 }}, pages 3664--3673, 2020.

\bibitem[Tu et~al.(2021)Tu, Yu, Wang, Birkbeck, Adsumilli, and Bovik]{RAPIQUE}
Zhengzhong Tu, Xiangxu Yu, Yilin Wang, Neil Birkbeck, Balu Adsumilli, and Alan~C Bovik.
\newblock { RAPIQUE: Rapid and accurate video quality prediction of user generated content }.
\newblock \emph{{IEEE Open Journal of Signal Processing}}, 2:\penalty0 425--440, 2021.

\bibitem[Tu et~al.(2022)Tu, Talebi, Zhang, Yang, Milanfar, Bovik, and Li]{maxvit}
Zhengzhong Tu, Hossein Talebi, Han Zhang, Feng Yang, Peyman Milanfar, Alan Bovik, and Yinxiao Li.
\newblock {Maxvit: Multi-axis vision transformer}.
\newblock \emph{{European Conference on Computer Vision}}, pages 459--479, 2022.

\bibitem[Wang et~al.(2020)Wang, Wang, Zhang, Owens, and Efros]{CNN-Spot}
Sheng-Yu Wang, Oliver Wang, Richard Zhang, Andrew Owens, and Alexei~A Efros.
\newblock {{CNN}-generated images are surprisingly easy to spot... for now}.
\newblock \emph{{IEEE/CVF Conference on Computer Vision and Pattern Recognition}}, pages 8695--8704, 2020.

\bibitem[Wang et~al.(2004)Wang, Bovik, Sheikh, and Simoncelli]{SSIM}
Zhou Wang, Alan~C Bovik, Hamid~R Sheikh, and Eero~P Simoncelli.
\newblock {Image quality assessment: from error visibility to structural similarity}.
\newblock \emph{{IEEE Transactions on Image Processing}}, 13\penalty0 (4):\penalty0 600--612, 2004.

\bibitem[Wang et~al.(2023)Wang, Bao, Zhou, Wang, Hu, Chen, and Li]{DIRE}
Zhendong Wang, Jianmin Bao, Wengang Zhou, Weilun Wang, Hezhen Hu, Hong Chen, and Houqiang Li.
\newblock {DIRE for Diffusion-Generated Image Detection}.
\newblock \emph{{arXiv preprint arXiv:2303.09295}}, 2023.

\bibitem[Yang et~al.(2021)Yang, Xiao, Li, Lan, and Wang]{Detecting-fake-images-by-identifying-potential-texture-difference}
Jiachen Yang, Shuai Xiao, Aiyun Li, Guipeng Lan, and Huihui Wang.
\newblock {Detecting fake images by identifying potential texture difference}.
\newblock \emph{{Future Gener. Comput. Syst.}}, 125:\penalty0 127--135, 2021.

\bibitem[Ying et~al.(2020)Ying, Niu, Gupta, Mahajan, Ghadiyaram, and Bovik]{PaQ2PiQ}
Zhenqiang Ying, Haoran Niu, Praful Gupta, Dhruv Mahajan, Deepti Ghadiyaram, and Alan Bovik.
\newblock { From Patches to Pictures (PaQ-2-PiQ): Mapping the Perceptual Space of Picture Quality }.
\newblock \emph{{ IEEE/CVF Conference on Computer Vision and Pattern Recognition (CVPR) }}, 2020.

\bibitem[Yu et~al.(2019)Yu, Davis, and Fritz]{yu2019attributing}
Ning Yu, Larry~S Davis, and Mario Fritz.
\newblock {Attributing Fake Images to GANs: Learning and Analyzing GAN Fingerprints}.
\newblock \emph{{IEEE/CVF International Conference on Computer Vision}}, pages 7556--7566, 2019.

\bibitem[Zeng et~al.(2017)Zeng, Zhang, and Bovik]{PQR}
Hui Zeng, Lei Zhang, and Alan~C Bovik.
\newblock { A probabilistic quality representation approach to deep blind image quality prediction }.
\newblock \emph{{arXiv preprint arXiv:1708.08190}}, 2017.

\bibitem[Zhang et~al.(2018)Zhang, Ma, Yan, Deng, and Wang]{DB-CNN}
Weixia Zhang, Kede Ma, Jia Yan, Dexiang Deng, and Zhou Wang.
\newblock { Blind Image Quality Assessment Using a Deep Bilinear Convolutional Neural Network }.
\newblock \emph{{IEEE Transactions on Circuits and Systems for Video Technology}}, 30\penalty0 (1):\penalty0 36--47, 2018.

\bibitem[Zhang et~al.(2019{\natexlab{a}})Zhang, Karaman, and Chang]{Freq-Spec}
Xu Zhang, Svebor Karaman, and Shih{-}Fu Chang.
\newblock {Detecting and Simulating Artifacts in GAN Fake Images}.
\newblock \emph{{ {IEEE} International Workshop on Information Forensics and Security, {WIFS} 2019, Delft, The Netherlands, December 9-12, 2019 }}, pages 1--6, 2019{\natexlab{a}}.

\bibitem[Zhang et~al.(2019{\natexlab{b}})Zhang, Karaman, and Chang]{Detecting-and-Simulating-Artifacts-in-GAN-Fake-Images}
Xu Zhang, Svebor Karaman, and Shih-Fu Chang.
\newblock {Detecting and Simulating Artifacts in GAN Fake Images}.
\newblock \emph{{IEEE International Workshop on Information Forensics and Security (WIFS) }}, pages 1--6, 2019{\natexlab{b}}.

\bibitem[Zhong et~al.(2024)Zhong, Xu, Li, Qian, and Zhang]{zhong2024patchcraft}
Nan Zhong, Yiran Xu, Sheng Li, Zhenxing Qian, and Xinpeng Zhang.
\newblock { Patchcraft: Exploring texture patch for efficient ai-generated image detection }.
\newblock \emph{{arXiv preprint arXiv:2311.12397}}, pages 1--18, 2024.

\bibitem[Zhu et~al.(2024)Zhu, Chen, Yan, Huang, Lin, Li, Tu, Hu, Hu, and Wang]{GenImage}
Mingjian Zhu, Hanting Chen, Qiangyu Yan, Xudong Huang, Guanyu Lin, Wei Li, Zhijun Tu, Hailin Hu, Jie Hu, and Yunhe Wang.
\newblock {LSUN: Construction of a Large-scale Image Dataset using Deep Learning with Humans in the Loop}.
\newblock \emph{{Advances in Neural Information Processing Systems}}, 36, 2024.

\end{thebibliography}
}


\end{document}